\PassOptionsToPackage{
    colorlinks=true,
    linkcolor=NavyBlue,
    citecolor=NavyBlue,
    urlcolor=NavyBlue,
    pagebackref,
    bookmarks
}{hyperref}

\PassOptionsToPackage{
    dvipsnames,
    svgnames
}{xcolor}

\documentclass[electronic]{vgtc}             

\graphicspath{{./figs/}}

\usepackage{times}

\usepackage{tabu}                      
\usepackage{booktabs}                  
\usepackage{cite} 
\usepackage{amssymb}
\usepackage{amsmath}

\newcommand{\para}[1]{\noindent{\textbf{#1}}}
\usepackage{enumitem}
\setlist[enumerate]{nosep, topsep=0pt, partopsep=0pt}
\setlist[itemize]{noitemsep, topsep=0pt, partopsep=0pt, leftmargin=*}

\usepackage{makecell}
\newcommand{\CellWithForceBreak}[2][c]
{\begin{tabular}[#1]{@{}c@{}}#2\end{tabular}}

\newcommand{\emax}{{e_\mathrm{max}}}
\newcommand{\eavg}{{e_\mathrm{avg}}}

\onlineid{1001}

\vgtccategory{Research}

\vgtcinsertpkg

\title{Extracting Complex Topology from Multivariate Functional Approximation: Contours, Jacobi Sets, and Ridge-Valley Graphs}

\author{Guanqun Ma\thanks{e-mail: guanqun.ma@utah.edu} \\
\scriptsize University of Utah
\and David Lenz \thanks{e-mail: dlenz@anl.gov} \\
\scriptsize Argonne National Laboratory
\and Hanqi Guo \thanks{e-mail: guo.2154@osu.edu} \\
\scriptsize Ohio State University
\and Tom Peterka \thanks{e-mail: tpeterka@mcs.anl.gov} \\
\scriptsize Argonne National Laboratory
\and Bei Wang \thanks{\vspace{-40pt} e-mail: beiwang@sci.utah.edu} \\
\scriptsize University of Utah}

\teaser{
\vspace{-1mm}
\includegraphics[width=0.8\linewidth]{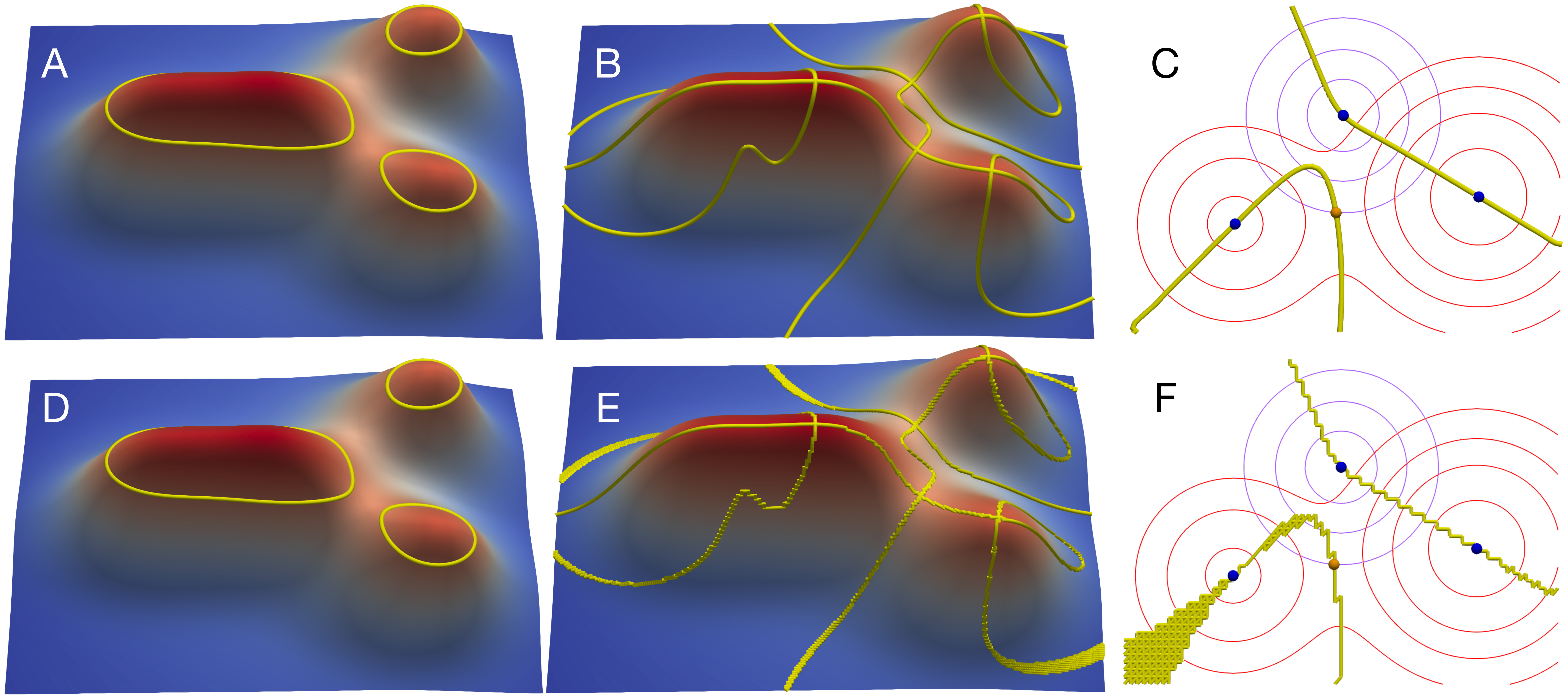}
\vspace{-3mm}
\centering
\caption{Extracting complex topological descriptors (highlighted in yellow) from MFA models. First, given an MFA model as input, we extract contours at a fixed isovalue in (A) and a ridge-valley graph in (B) directly from the MFA model, without discretizing the entire domain. Next, for a pair of MFA models, we extract the Jacobi set in (C), where the contours from each model are shown in purple and red, respectively. Finally, we compare these topological descriptors with the contour (D), ridge-valley graph (E), and Jacobi set (F) derived from discrete representations of the MFA models. Local maxima and saddles  are indicated by blue and orange dots in (C) and (F), respectively.}
\label{fig:teaser-image}
}

\abstract{
Implicit continuous models, such as functional models and implicit neural networks, are an increasingly popular method for replacing discrete data representations with continuous, high-order, and differentiable surrogates. These models offer new perspectives on the storage, transfer, and analysis of scientific data. In this paper, we introduce the first framework to directly extract complex topological features---contours, Jacobi sets, and ridge-valley graphs---from a type of continuous implicit model known as multivariate functional approximation (MFA). MFA replaces discrete data with continuous piecewise smooth functions. Given an MFA model as the input, our approach enables direct extraction of complex topological features from the model, without reverting to a discrete representation of the model. 
Our work is easily generalizable to any continuous implicit model that supports the queries of function values and high-order derivatives. Our work establishes the building blocks for performing topological data analysis and visualization on implicit continuous models.

}


\begin{document}

\maketitle

\section{Introduction} 
\label{sec:introduction}

Implicit continuous models---functional models (e.g., spline representations and radial basis functions) and implicit neural networks (e.g., sinusoidal representation networks \cite{sitzmann2020implicit})---have gained growing interest in scientific data analysis and visualization. These implicit models replace discrete data representations with continuous, high-order, and differentiable data representations, thus improving the flexibility and efficiency of scientific workflows~\cite{majdisova2017radial, sitzmann2020implicit,mildenhall2020nerf} and enabling complex analysis of scientific data~\cite{niemeyer2021giraffe,majdisova2017radial,sitzmann2019scene}. 

A notable example of an implicit continuous model is the multivariate functional approximation (MFA)~\cite{peterka2018foundations,peterka2022multivariate}. 
An MFA model represents discrete data by approximating it with a set of piecewise smooth polynomial functions. It allows for the evaluation of both function values and high-order derivatives at any point within the domain. By utilizing a mesh-free, uniform representation, MFA enables the remapping between different data representations and supports spatiotemporal analysis in a continuous setting. Supported by the SciDAC RAPIDS Institute \cite{rapids2} of the U.S. Department of Energy (DOE), MFA has found applications in high-energy physics~\cite{HepOnHPC} and climate science~\cite{seahorce}. It has proven effective as an intermediate data representation for high-quality volume rendering~\cite{sun2023scalable}.

Despite recent advancements, we are only beginning to explore the potential of implicit continuous models in supporting topological data analysis and visualization. Topological descriptors have been instrumental in enabling a number of scientific visualization tasks, such as feature detection, tracking, and clustering; see~\cite{heine2016survey,yan2021scalar} for surveys. However, existing topological methods work with data represented on discrete grids. Naively applying these methods to implicit continuous models requires discretization of the entire domain. However, this discretization effectively reduces the high-order model to a set of low-order features, which can introduce aliasing artifacts.

We therefore tackle a key question: Given a continuous model as the input, such as an MFA, how can we extract topological descriptors \emph{directly} from the model without resorting to discretization of the entire domain? Recently, Ma et al.~\cite{ma2024critical} extracted critical points directly from an MFA model without discretization. Building on~\cite{ma2024critical}, we present a method for directly extracting more complex features from MFA models: contours, Jacobi sets and ridge-valley graphs.

Topological descriptors such as contours, ridge-valley graphs, and Jacobi sets are important in studying scientific data. 
Given a scalar function, a contour (or level set) represents a part of its  domain that remains a constant value. 
Contours thus provide insights into the structural information of data at user-specified thresholds~\cite{lorensen1987marching}. 
A ridge-valley graph delineates ridges and valleys of a scalar function that are invariant under translations, rotations, uniform magnification, and monotonic transformations~\cite{gregory2013ridge}.  
The Jacobi set, on the other hand, arises from multiparameter data analysis: given a pair of scalar functions, it identifies points where their gradients are aligned. In particular, the Jacobi set reveals the relationship between a pair of functions by analyzing the critical points of one function restricted to the contour of another~\cite{edelsbrunner2002jacobi}. 
 
The framework presented in this paper takes full advantage of the continuous and differentiable representations afforded by MFA models. 
Our contributions are summarized below.  
\begin{itemize}
\item We present a framework for extracting contours directly from continuous models---specifically, MFA models---without discretizing the entire domain. We utilize particle tracing along directions perpendicular to the gradient to accurately locate contours, by leveraging the ability of these models to query function values and derivatives at any point in the domain. 
\item We extract Jacobi sets and ridge-valley graphs from MFA models by converting the problem to contour extraction from derived functions. This approach simplifies and unifies the extraction process across multiple topological descriptors. 
\end{itemize}

\section{Related Work}
\label{sec:related-work}

\para{Functional models and implicit neural networks.}
Implicit continuous models represent scalar functions in a smooth and differentiable manner~\cite{mello2024neural,bloomenthal1997introduction}. Unlike discrete representations that may suffer from limited resolution, continuous models can be evaluated at any point in the domain. 
They support queries of function values and high-order derivatives, which are essential for data analysis and visualization~\cite{gomes2009implicit, ma2024critical}.

Implicit neural networks---a family of continuous models---have gained significant attention in scientific machine learning~\cite{sitzmann2020implicit,park2019deepsdf,yu2022monosdf}. These models enable the representation of complex shapes, scenes, and signals without explicit discretization.
One prominent example of an implicit neural network is the sinusoidal representation network (SIREN)~\cite{sitzmann2020implicit}. SIREN utilizes periodic activation functions for implicit neural representations (INR) and has been used in volumetric data compression~\cite{lu2021compressive}. 

\para{Multivariate functional approximation.}~MFA has emerged as a powerful method for representing scientific datasets using continuous B-spline functions (see e.g.,~\cite{peterka2018foundations, peterka2022multivariate, lenz2023customizable}). An MFA model could be considered as a form of scattered data approximation (SDA) that builds continuous functions to approximate spatial datasets~\cite{wendland2004scattered}. Several SDA methods have been proposed with functional approximations based on splines~\cite{deBoor2001guide}, wavelets~\cite{jansen2005second}, and radial basis functions~\cite{majdisova2017radial}.  

An MFA model distinguishes itself from other SDA methods by leveraging the properties of B-splines to achieve high-order continuity and efficient computation. 
It reduces storage due to its compact representation and supports the computation of function values and derivatives of any order at any point in the domain~\cite{lenz2023customizable}. Lenz et al.~\cite{lenz2023customizable} used MFA to enable customizable approximations of complex datasets. Sun et al. \cite{sun2023scalable,sun2024mfa} developed scalable volume visualization techniques based on MFA representations, demonstrating high-quality rendering with reduced computational overhead. 

Ma et al.~\cite{ma2024critical} introduced critical point extraction from an MFA model. As critical points are the simplest forms of topological features, their work could be considered as a first step toward extracting complex topology from MFA. 

\para{Contour extraction} is a fundamental task in scientific visualization, allowing the exploration of complex functions by identifying curves or surfaces along which the function matches a user-specified constant~\cite{lorensen1987marching}. 
Classic methods use the marching cubes algorithm~\cite{lorensen1987marching} and its variants~\cite{nielson2003marching,thomas2003efficient} to extract contours from discrete data on grids. 
For 3D (resp.~2D) data, the marching cubes (resp.~marching squares) algorithm examines how each contour passes through a cube/square. 
Ju et al.~\cite{ju2002dual} presented a dual contouring algorithm to extract contours from a signed grid, utilizing gradient information to improve the extraction quality.  
While classic methods are based on discrete data, continuous implicit models give accurate evaluation of function values and derivatives. As shown in this paper, utilizing implicit models could bypass the data discretization step and improve accuracy. 

Our approach is inspired by methods used in surface–surface intersection for CAD models, with the distinction that one of the surfaces is derived from the gradient of the other. It also relates to the literature on non-isolated root finding. For example, Dokken~\cite{dokken1985Bsplines} employed recursive subdivision techniques to compute intersections between a B-spline surface and a plane. While his method is tailored to B-splines, our approach is adaptable to more general classes of functions. Polynomial homotopy continuation~\cite{allgower2012numerical} provides an alternative framework for solving systems of equations. Theisel et al.~\cite{theisel2004stream} tracked critical points in a 2D time-dependent vector field by integrating streamlines of a derived field. In a manner similar to our method, they used a Runge–Kutta integration scheme to ensure that the resulting critical lines are independent of an underlying grid. 

\para{Jacobi sets} capture the relationships among multiple scalar functions by identifying points where their gradients are aligned~\cite{edelsbrunner2002jacobi}. 
Edelsbrunner and Harer~\cite{edelsbrunner2002jacobi} formalized the concept of Jacobi sets for multiple Morse functions, exploring the topological properties and applications. Jacobi sets and Reeb spaces are commonly used topological descriptors for multiparameter data analysis. 
A Reeb space captures the structure of a multiparameter mapping by compressing the connected components of its level sets. 
Jacobi sets and Reeb spaces are connected through a mapping between their singularities~\cite{carr2014joint}. 
Such a mapping was studied by Chattopadhyay et al. \cite{chattopadhyay2014extracting}.  
Kl\"otzl et al. utilized a local bilinear method to approximate a Jacobi set with good accuracy~\cite{klotzl2022local}, and proposed a topological connection method for Jacobi set computation \cite{klotzl2022reduced}, improving visual clarity while preserving topological structure.

Tierny and Carr \cite{tierny2017data} built on Jacobi set to compute the Reeb space of a bivariate function defined on a tetrahedral mesh.
Sharma and Natarajan \cite{sharma2022jacobi} used Jacobi sets to identify fiber surfaces. Meduri et al.~\cite{meduri2024jacobi} proposed Jacobi set simplification for tracking topological features in time-varying scalar functions.

A closely related concept is the Pareto set, which studies scalar functions based on consensus or disagreement among their critical points, ascending and descending manifolds, and connectivity~\cite{huettenberger2013towards}. 
Huettenberger et al.~\cite{huettenberger2015comparison} explored the relationship between Pareto sets and Jacobi sets, establishing subset and equivalence relations between them. 

Compared with discrete representations, continuous implicit models enable precise identification of Jacobi sets without artifacts from data discretizations. Leveraging the continuous and differentiable nature of MFA models could improve the efficiency and accuracy of Jacobi set extraction, benefiting multiparameter data analysis and visualization.

\para{Ridge-valley graphs.}~Damon~\cite{damon1998generic} proposed ridge-valley-connector-curves that describe the global structure of height ridges. A point belongs to a height ridge if the scalar function has a local maximum in the direction orthogonal to its gradient. 
Norgard and Bremer introduced the ridge-valley graph~\cite{gregory2013ridge} based on the notion of Jacobi ridges, which behave similarly to the curves proposed in~\cite{damon1998generic}. 
Jacobi ridges are points where the gradient magnitude has a local minimum along the level set~\cite{gregory2013ridge}. 
However, unlike height ridges~\cite{eberly1994ridges}, Jacobi ridges are invariant under monotonic transformations, although their structures are similar and their formulations coincide for quadratic functions~\cite{gregory2013ridge}.
A ridge-valley graph is the Jacobi set of a function and its squared gradient magnitude (see~\cref{sec:technical-background}).

Later studies introduced different methods for extracting ridges. For instance, Anisotropic Gaussian Kernel (AGK) is employed to enhance sensitivity and robustness~\cite{lopez2015unsupervised}. 
Reisenhofer and King \cite{reisenhofer2019edge} refined local features by integrating the contrast-invariant phase congruency measure with $\alpha$-molecules. 
\section{Technical Background}
\label{sec:technical-background}
We provide the background of MFA, contours, Jacobi sets, and ridge-valley graphs to support the understanding of our method.

\subsection{Multivariate Functional Approximation}
\label{sec:MFA-concept}
MFA models are tensor-product B-spline functions that approximate a discrete dataset. We briefly review B-splines below; see~\cite{deBoor2001guide,piegl1997nurbs} for a detailed presentation of B-spline theory. 

\begin{figure}[!ht]
\vspace{-5mm}
\centering
\includegraphics[width=.8\linewidth]{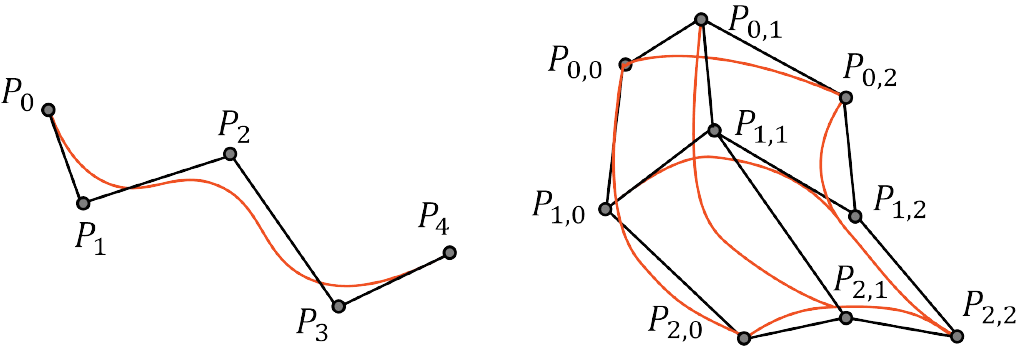}
\vspace{-2mm}
\caption{A 1D B-spline curve (left) and a 2D B-spline surface (right).
$P_i$ and $P_{i,j}$ are control points; control meshes are in black; approximated curves or surfaces are in red.}
\label{fig:mfa-1d-2d}
\vspace{-2mm}
\end{figure}

A degree $p$ B-spline curve is a piecewise-polynomial function. It has $p-1$ continuous derivatives and is composed of degree $p$ polynomial pieces. 
The junction points between these pieces are referred to as \emph{knots}, and the intervals between knots are \emph{knot spans} or simply \emph{spans}. The shape of the curve is determined by a set of \emph{control points} distributed across the domain. Although the curve loosely follows a polyline formed by these control points, it does not necessarily pass through them; see~\cref{fig:mfa-1d-2d}.  

Mathematically, a B-spline $F$ is defined as a combination of \emph{basis functions} $N_j$, weighted by the control point positions $P_j$:
\begin{equation}
    F(u) = \sum_{j=0}^{n-1} N_j(u) P_j.
\end{equation}
In 1D, given a set of points located at $\{u_0, \ldots, u_{m-1}\} \subset [0,1]$, where each $u_i$ is associated with a value $f_i$, the best-fit B-spline curve has optimal control points to minimize the pointwise error:
\begin{equation}
    \min_P \left(\frac{1}{m}\sum_{i=0}^{m-1} |f_i - \sum_{j=0}^n N_j(u_i) P_j|^2 \right)^{1/2}.
\end{equation}
 
In this paper, we focus on MFA in the 2D case. A 2D B-spline surface of degree $p$ is defined using two sets of knots, $\{t^{(1)}_{j_1}\}_{j_1=0}^{n_1+p}$ and $\{t^{(2)}_{j_2}\}_{j_2=0}^{n_2+p}$, corresponding to the $u$ and $v$ dimensions, respectively. 
$\{P_{j_1,j_2}\}_{j_1,j_2=0}^{n_1-1,n_2-1}$ is the set of control points.
The 2D basis functions are constructed as the product of 1D basis functions in each dimension: $N_{j_1,j_2}(u,v) = N^{(1)}_{j_1,p}(u)N^{(2)}_{j_2,p}(v)$, where $N^{(1)}_{j_1,p}(u)$ and $N^{(2)}_{j_2,p}(v)$ are 1D basis functions for the $u$ and $v$ dimensions. 
The resulting 2D B-spline surface is expressed as:
\begin{equation}
\label{eq:B-spline-2d}
    F(u,v) = \sum_{j_1 = 0}^{n_1-1} \sum_{j_2 = 0}^{n_2-1}N^{(1)}_{j_1,p}(u)N^{(2)}_{j_2,p}(v) P_{j_1,j_2}.
\end{equation}
Finally, a 2D span is simply the tensor product of 1D spans. For example, a 2D span corresponds to a region: 
\[[t^{(1)}_{j_1}, t^{(1)}_{j_1+1}]\times[t^{(2)}_{j_2}, t^{(2)}_{j_2+1}].\]

\subsection{Critical Point Extraction From an MFA Model}
\label{sec:CPE}
Since an MFA model is a piecewise-polynomial function, critical points can be directly extracted from an MFA model following the approach outlined by Ma et al.~\cite{ma2024critical}:
\begin{enumerate}[noitemsep,leftmargin=*]
    \item Span filtration: Each span is constrained within the convex hull of its corresponding control points. By analyzing the control points of the first derivatives, we identify and exclude spans that are impossible to contain critical points, thereby reducing the spans we actually work with.
    \item Critical point extraction: For the remaining spans, we use Newton's method to compute critical points in each span. The iterative update formula is given by:
    \begin{equation}
      \mathbf{x}_{n+1}=\mathbf{x}_n-H(\mathbf{x}_n)^{-1}\nabla f(\mathbf{x}_n),
    \end{equation} 
    where $H(\mathbf{x}_n)$ is the Hessian matrix and $\nabla f(\mathbf{x}_n)$ is the gradient of the MFA model $f$ evaluated at $\mathbf{x}_n$. Since MFAs are differentiable functions, each of these queries is exact.
    \item Deduplication: To ensure uniqueness, we remove duplicated critical points using spatial hashing.
\end{enumerate}
According to \cite{ma2024critical}, the overall time complexity in a 2D domain is $\mathcal{O}(i_{max}p^{4}n)$, where $i_{max}$ is the maximum number of iterations using Newton’s method and $n$ is the number of spans.

\subsection{Particle Tracing}
\label{sec:particle-tracing}
Particle tracing involves finding the trajectories of particles moving through a vector field, essential in visualizing flow patterns such as streamlines and pathlines~\cite{peterka2011study}. 
The motion of a particle can be described by an initial value problem (IVP) in the form of an ordinary differential equation:
\begin{equation}
\label{eq:initial-value-problem}
    \frac{d\mathbf{x}}{dt} = \mathbf{v}(t,\mathbf{x}), \mathbf{x}(t_0)=\mathbf{x}_0,
\end{equation}
where $\mathbf{x}(t)$ represents the position of a particle at time $t$ and $\mathbf{v}(t,\mathbf{x})$ is the velocity at position $\mathbf{x}$ and time $t$. 
Particle tracing is performed by numerically solving the IVP~\cite{pokrajac2002efficient} in~\eqref{eq:initial-value-problem}. 
To do so, numerical integration methods such as the Runge-Kutta method are used. 

The classic Runge-Kutta method (RK4)~\cite{Press2007numerical} is a fourth-order method to approximate $\mathbf{x}_{n+1}$ from $\mathbf{x}_{n}$ in~\eqref{eq:initial-value-problem}:
\begin{align}
\mathbf{x}_{n+1} & =\mathbf{x}_n+\frac{s}{6}\left(\mathbf{k}_1+2 \mathbf{k}_2+2 \mathbf{k}_3+\mathbf{k}_4\right), \\
t_{n+1} & =t_n+s,
\end{align}
where
$\mathbf{k}_1 =\mathbf{v}\left(t_n, \mathbf{x}_n\right)$, 
$\mathbf{k}_2 =\mathbf{v}\left(t_n+\frac{s}{2}, \mathbf{x}_n+\frac{s}{2}\mathbf{k}_1\right)$, 
$\mathbf{k}_3 =\mathbf{v}\left(t_n+\frac{s}{2}, \mathbf{x}_n+\frac{s}{2}\mathbf{k}_2\right)$, 
and $\mathbf{k}_4 =\mathbf{v}\left(t_n+s, \mathbf{x}_n+s \mathbf{k}_3\right)$, and $s>0$ is the step size. RK4 yields $\mathbf{x}_{n+1}$ as an approximation to $\mathbf{x}$ at time $t_{n+1}$. The local truncation error per step is $O(s^5)$ and the total accumulated error is $O(s^4)$ \cite{Press2007numerical}.

\subsection{Jacobi Set}
Studying natural phenomena often involves analyzing the relationship between two functions defined on the same domain, such as temperature distribution within a layer of constant salinity \cite{edelsbrunner2002jacobi}. Jacobi sets capture this relationship by analyzing the critical points of one function restricted to the contours of the other.

A function is a Morse function if all its critical points are non-degenerate and have distinct function values. Given two Morse functions $f,g: \mathbb{M}\rightarrow \mathbb{R}$ defined on a manifold $\mathbb{M}$, the Jacobi set $\mathbb{J} = \mathbb{J}(f,g)=\mathbb{J}(g,f)$ is defined as the closure of the set of points where the gradients of $f$ and $g$ are linearly dependent \cite{edelsbrunner2002jacobi}: 
\begin{equation}
\label{eq:jacobi-set}
\begin{aligned}
     \mathbb{J}(f,g)=\operatorname{cl}\{\mathbf{x} \in \mathbb{M} \mid \nabla f(\mathbf{x})+\lambda \nabla g(\mathbf{x})=0 \text { or } \\ \nabla g(\mathbf{x})+\lambda \nabla f(\mathbf{x})=0\}, 
\end{aligned}
\end{equation}
for some $\lambda \in \mathbb{R}$. This condition implies that
\begin{equation}
     \mathbb{J}(f,g)=\operatorname{cl}\{\mathbf{x} \in \mathbb{M} \mid \mathbf{x}  \text{ is a critical point of } f +\lambda g \text { or } g+\lambda f\}, 
\end{equation}
for some $\lambda \in \mathbb{R}$~\cite{edelsbrunner2002jacobi}. Alternatively, let $g^{-1}(t)$ denote the level set of $g$ at $t\in\mathbb{R}$, and let $f_t:g^{-1}(k)\rightarrow \mathbb{R}$ be the restriction of $f$ on $g^{-1}(t)$, then $\mathbb{J}(f,g)$ can equivalently be defined \cite{harsh2015local} as
\begin{equation}
    \mathbb{J}(f,g)=\operatorname{cl}\{\mathbf{x} \in \mathbb{M} \mid \mathbf{x} \text{ is a critical point of } f_t\}.
\end{equation}
The critical points of $f_t$ are known as the \emph{restricted critical points} of $f$ with respect to $g$. Symmetrically, we can define restricted critical points of $g$ with respect to $f$.

The Jacobi set $\mathbb{J}(f,g)$ is an embedded 1-manifold in $\mathbb{M}$ (i.e, a set of curves)~\cite{edelsbrunner2002jacobi}. It includes all critical points of $f$ and $g$. The function $f_t$ is a Morse function except at a finite set of measure zero, where $f_t$ changes in criticality at critical points of $g$~\cite{harsh2015local}. The alignment of restricted critical points changes at critical points of $f$ and $g$ \cite{harsh2015local}. Examples of Jacobi sets are illustrated in~\cref{fig:teaser-image} (C) and (F). 
The Jacobi set $\mathbb{J}(f,g)$ is the zero level set of the comparison measure $\kappa_x=||\nabla f(x)\times \nabla g(x)||$ \cite{simplification2011suthambhara}.

\subsection{Ridge Valley Graph}
\label{sec:ridge-valley-graph}
Ridges and valleys are highly sought-after features across diverse fields, from image processing \cite{shokouh2021ridge} to combustion simulations \cite{reisenhofer2016shearlet}.
A ridge-valley graph is a complete description of all ridges and valleys in a scalar function  $f:\mathbb{M}\rightarrow\mathbb{R}$~\cite{gregory2013ridge}. It refers to Jacobi ridges that satisfy all five desired invariants~\cite{gregory2013ridge}. A ridge-valley graph of $f$ corresponds to a non-generic Jacobi set $\mathbb{J}(f,||\nabla f||^2)$ formed by $f$ and its squared gradient magnitude~\cite{gregory2013ridge}. Following~\cite{gregory2013ridge}, a ridge-valley graph of $f$ consists of nodes and arcs with the following properties (see~\cref{fig:teaser-image} (A)):
\begin{enumerate}
\item Arcs: Smoothly embedded curves in $\mathbb{M}$.
\item Nodes: Valence-4 nodes correspond to the critical points of $f$. Valence-2 nodes are the points where classification changes (see below).
\item Classification: An arc is consistently classified as a (pseudo-) ridge or a (pseudo-) valley. The classification depends on the behavior of $||\nabla f||^2$ along the contours of $f$ and the behavior of $f$ in the direction tangent to these contours. 
A ridge (resp. valley) point is one where $||\nabla f||^2$ is minimal along a contour and $f$ is maximal (resp. minimal) tangent to the contour. 
A pseudo-ridge (resp. pseudo-valley) point is one where $||\nabla f||^2$ is maximal along a contour and $f$ is maximal (resp. minimal) tangent to the contour.
\end{enumerate}
A ridge-valley graph is invariant under translations, rotations, and uniform magnification in the spatial variables. Additionally, it is defined locally and remains invariant under monotonic transformations of $f$ \cite{gregory2013ridge}. 
Examples of ridge-valley graphs are illustrated in~\cref{fig:teaser-image} (B) and (E). 

\section{Method}
\label{sec:method}

To the best of our knowledge, we present the first framework for extracting contours, Jacobi sets, and ridge-valley graphs from continuous implicit models. 
We demonstrate our framework using MFA models; however, it is broadly applicable and can be adapted to any implicit model that allows querying of function values and higher-order derivatives.

\subsection{Contour Extraction}
\label{sec:contour-extraction}

Given a Morse function $f: \mathbb{M}\rightarrow\mathbb{R}$ on a 2-manifold $\mathbb{M}$, a \emph{contour} (level set) $\gamma_a$ of $f$ defined at a given threshold (isovalue) $a \in \mathbb{R}$ is given by $\gamma=\gamma_a:=\{\mathbf{x}\in\mathbb{M} \mid f(\mathbf{x})=a\}$. 
At any point $\mathbf{x}=(x_1,x_2)^\top \in \gamma$, the gradient is orthogonal to $\gamma$, meaning the tangent direction $\mathbf{m}$ of $\gamma$ at $\mathbf{x}$ is perpendicular to $\nabla f(\mathbf{x})$. Let $\nabla f=(f_{x_1},f_{x_2})^\top$ represent the gradient, then 
\begin{equation}
\label{eq:direction}
    \mathbf{m}=\frac{1}{||\nabla f||}(-f_{x_2},f_{x_1})^{\top},
\end{equation}
where $||\nabla f||=\sqrt{f_{x_1}^2+f_{x_2}^2}$ is the gradient magnitude.

Contour extraction can be formulated as a particle tracing problem, where a particle moves along the contour by following the tangent direction $\mathbf{m}$. To numerically integrate this trajectory, we employ the RK4 method by setting $\mathbf{v}=\mathbf{m}$ and ignoring time  (see~\cref{sec:particle-tracing}). Let $s>0$ be the step size. 
\begin{align}
\label{eq:rk4}
\mathbf{x}_{n+1} & =\mathbf{x}_n+\frac{s}{6}\left(\mathbf{k}_1+2 \mathbf{k}_2+2 \mathbf{k}_3+\mathbf{k}_4\right), 
\end{align}
where
$\mathbf{k}_1 =\mathbf{m}\left(\mathbf{x}_n\right)$, 
$\mathbf{k}_2 =\mathbf{m}\left(\mathbf{x}_n+\frac{s}{2}\mathbf{k}_1\right)$, 
$\mathbf{k}_3 =\mathbf{m}\left(\mathbf{x}_n+\frac{s}{2}\mathbf{k}_2\right)$, and 
$\mathbf{k}_4 =\mathbf{m}\left(\mathbf{x}_n+s \mathbf{k}_3\right)$. 
Starting from an initial point $\mathbf{p}_0$ on the contour $\gamma$, we obtain a sequence of points $\{\mathbf{x}_0, \mathbf{x}_1, \mathbf{x}_2, \cdots\}$ by iteratively applying the RK4 method, thereby tracing the contour.

Our contour extraction method comprises three main steps:
\begin{enumerate}
    \item Initialization: Identify a set of starting points on the contour.
    \item Particle tracing: Utilize the RK4 method to trace the contour from each starting point.
    \item Connecting: Incorporate critical points and connect the traced points to form continuous segments of the contour.
\end{enumerate}
We discuss these steps in detail in~\cref{sec:initialization-step,sec:particle-tracing-step,sec:connecting-step}. 

\subsubsection{Initialization}
\label{sec:initialization-step}
To find points satisfying $f(\mathbf{x})=a$ for a given $a$, we convert the problem into a root-finding problem. We define $f_a(\mathbf{x}):=f(\mathbf{x})-a$ and search for the roots of $f_a$ by setting $f_a = 0$. We use a normalized gradient descent method to solve the root-finding problem, where the iteration formula is 
\begin{equation}
\label{eq:gradient_descent}
    \mathbf{x}_{n+1} = \mathbf{x}_n - \alpha\frac{\nabla f_a(\mathbf{x}_{n})}{||\nabla f_a(\mathbf{x}_n)||},
\end{equation}
with an adaptive step size $\alpha =  \frac{f_a(\mathbf{x}_n)}{||\nabla f_a(\mathbf{x}_n)||}$. 

In the context of MFA, the domain is partitioned into spans, each representing a polynomial function. Within each span, we uniformly sample $(p+3)^d$ initial points, where $p$ is the polynomial degree and $d=2$ is the dimension. From each initial point, we apply the normalized gradient descent to find a starting point satisfying $f(\mathbf{x})=a$ (or equivalently, $f_a(\mathbf{x})=0$). To avoid redundancy, we remove starting points that are closer than $s$ to other starting points (duplication removal). The root-finding process is confined within the current span. Since each span is processed independently, this step is naturally adapted to multithreading.

\subsubsection{Particle Tracing}
\label{sec:particle-tracing-step}
In the particle tracing step, we continue to treat each span independently. 
Within a given span, we trace a piece of the contour (a trajectory) using RK4 method (see~\cref{eq:rk4}) from each starting point obtained from the initialization step (\cref{sec:initialization-step}). The procedure is as follows:
\begin{enumerate}
\item \textbf{Forward tracing}: Trace the trajectory forward following $\mathbf{m}$.
\item \textbf{Boundary handling}: If the trajectory exits within the current span, stop tracing. To make sure the distance between the last point inside the span and the first point in an adjacent span does not exceed $s$, we set $s \leftarrow \frac{s}{2}$ and process with RK4 one more time within the boundary. 
\item \textbf{Backward tracing}: To capture the entire trajectory passing through the starting point, if the trajectory does not form a loop, we also trace backward from the starting point in the opposite direction $-\mathbf{m}$. If the trajectory returns to the starting point (indicating that it forms a loop), backward tracing is unnecessary.
\end{enumerate}

When a trajectory reaches a critical point, the gradient magnitude of $f$ becomes negligible, and an RK4 update cannot proceed effectively. In such cases, we stop tracing when $||\nabla f(\mathbf{x}_n)||$ is smaller than a threshold. In practice, as the trajectory approaches a critical point, the step size $||\mathbf{x}_{n+1}-\mathbf{x}_n||$ decreases significantly. Therefore, in all the experiments, we stop tracing when $||\mathbf{x}_{n+1}-\mathbf{x}_{n}||<0.5s$. To prevent error accumulation during particle tracing, we utilize a correction step at every point. If $|f(\mathbf{x}_n)-a|>\epsilon$, where $\epsilon$ is the threshold to control the accuracy, we adjust the point position using the normalized gradient descent described in \cref{eq:gradient_descent}.

\begin{figure}[!ht]
\centering
\vspace{-3mm}
\includegraphics[width=.7\linewidth]{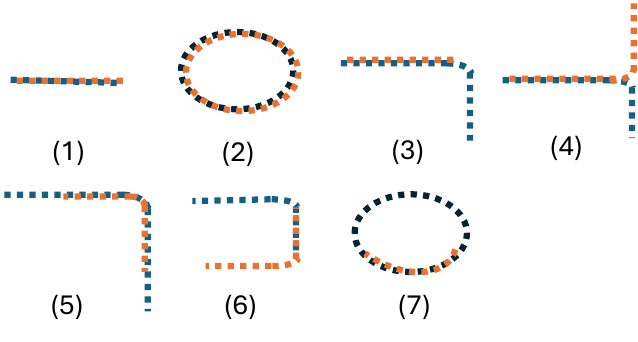}
\vspace{-5mm}
\caption{Illustrations of duplicated trajectories as candidates for duplication removal. Two trajectories that contain duplicated segments are shown in orange and blue respectively.}
\vspace{-4mm}
\label{fig:trajectory-duplication}
\end{figure}

Since multiple starting points may trace the same trajectory, duplication removal is necessary. Even when we stop tracing near critical points, it is possible for trajectories to skip over these critical points, leading to multiple recordings of the same trajectory. We deal with various duplication scenarios as illustrated in~\cref{fig:trajectory-duplication}. 
If the distance between points from two different trajectories is less than $s$, we consider these points as \emph{corresponding}. The key idea behind duplication removal is identifying duplicated segments by searching for corresponding endpoints of two trajectories. 

\subsubsection{Connecting}
\label{sec:connecting-step}
After removing duplications, we connect trajectories to form continuous contours. 
\begin{enumerate}[noitemsep]
\item \textbf{Inserting critical points}: Following~\cref{sec:CPE}, we identify and insert critical points $\mathbf{x}$ satisfying $||f(\mathbf{x})-t||<\epsilon$. These critical points can have valence $>2$. Since particle tracing alone cannot guarantee to find these points, we add them manually to ensure completeness.   
\item \textbf{Connecting to critical points}: For trajectories within a span, we 
connect their endpoints to nearby critical points if they are within a distance $\leq 2s$. 
On a contour, the valence of a critical point may be $\geq 2$, whereas that of a regular point is $2$, since the gradient is nonzero at a regular point. Consequently, if a trajectory contains more than one point, it can only connect to another trajectory or a critical point.
\item \textbf{Connecting trajectories across spans}: To avoid duplication, we only compare trajectories with those in adjacent spans with higher span indices. If endpoints from different spans are within a distance of $2s$, we connect them.
\item \textbf{Connecting trajectories within a span}: We connect endpoints of trajectories within the same span if their distance is $<2s$.
\item \textbf{Connecting within a trajectory}: Within each trajectory, we connect all points sequentially for continuity.
\end{enumerate}

\subsection{Jacobi Set Extraction}
Following~\cref{eq:jacobi-set}, a point $\mathbf{x}=(x_1,x_2)^{\top}$ belongs to the Jacobi set $\mathbb{J}:=\mathbb{J}(f,g)$ if one of the following conditions holds: $\nabla f(\mathbf{x}) = 0$, $\nabla g(\mathbf{x}) = 0$, or the gradients $\nabla f(\mathbf{x})$ and $\nabla g(\mathbf{x})$ are aligned (parallel). As noted in~\cite{simplification2011suthambhara,local2004edelsbrunner}, such points lie on the zero level set of $||\nabla f(\mathbf{x}) \times \nabla g(\mathbf{x})||$. We take inspiration from this formulation.
In a 2D domain, the gradients are given by $\nabla f = (f_{x_1}, f_{x_2})^{\top}$ and $\nabla g = (g_{x_1}, g_{x_2})^{\top}$. To utilize the cross product, we extend these gradients to 3D by adding a zero component: $\nabla f_3 = (f_{x_1}, f_{x_2}, 0)^{\top}$ and $\nabla g_3 = (g_{x_1}, g_{x_2},0)^{\top}$. To identify $\mathbf{x} \in \mathbb{J}$, the following cross product must be zero:
\begin{align}
\label{eq:cross}
\nabla f_3(\mathbf{x})\times \nabla g_3(\mathbf{x}) = (0,0,f_{x_1}g_{x_2}-f_{x_2}g_{x_1})^{\top} = (0,0,0)^{\top}.
\end{align}
This condition implies that either $\nabla f(\mathbf{x})=0$, $\nabla g(\mathbf{x})=0$, or $\nabla f(\mathbf{x})$ and $\nabla g(\mathbf{x})$ are parallel, which is consistent with the definition in ~\cref{eq:jacobi-set}. We then define
\begin{equation}
\label{eq:jaocbi-set-h}
    h=f_{x_1}g_{x_2}-f_{x_2}g_{x_1}. 
\end{equation}
Extracting the Jacobi set $\mathbb{J}$ becomes equivalent to finding the contour of $h$ with $a = 0$ (i.e., the 0-contour). 
We then apply the contour extraction described in~\cref{sec:contour-extraction} to determine the Jacobi set.

In MFA, $f$ and $g$ are constructed with the same span settings. Their only difference lies in their control points. As $f$ and $g$ share the same spans, $h$ is also a polynomial function within each span. Consequently, our Jacobi set extraction process can be seamlessly integrated with our contour extraction method that treats each span independently. 
In the final step, we extract the critical points of $h$. 

To utilize the contour extraction method, we require the gradient of $h$ given by:
\begin{equation}
\label{eq:h-gradient}
\nabla h = 
    \begin{pmatrix}
    h_{x_1} \\ h_{x_2}
\end{pmatrix}
=
\begin{pmatrix}
       f_{x_1}g_{x_1x_2}+g_{x_2}f_{x_1x_1} - f_{x_2}g_{x_1x_1} - g_{x_1}f_{x_1x_2} \\
    f_{x_1}g_{x_2x_2}+g_{x_2}f_{x_1x_2} - f_{x_2}g_{x_1x_2} - g_{x_1}f_{x_2x_2}
\end{pmatrix}.
\end{equation}
In practice, we extract the critical points $\mathbf{x}$ of $h$ that satisfy $||h(\mathbf{x})||<\epsilon$ and incorporate them into the result.

\subsection{Ridge-Valley Graph Extraction}
Following~\cref{sec:ridge-valley-graph}, a ridge-valley graph is a special type of Jacobi set $\mathbb{J}(f,||\nabla f||^2)$.  
Let $g=\|\nabla f\|^2$. The gradient of $g$ is given by $\nabla g = 2H \nabla f$, where $H$ is the Hessian matrix of $f$.

Formally, $H$ and $\nabla g$ are derived as follows:  
\begin{align}
H & = \begin{pmatrix} f_{x_1x_1} & f_{x_1x_2} \\ f_{x_1x_2} & f_{x_2x_2} 
\end{pmatrix},\\
 \nabla g & =
2H\nabla f =
2\begin{pmatrix}
   f_{x_1x_1}f_{x_1}+f_{x_1x_2}f_{x_2} \\f_{x_1x_2}f_{x_1}+f_{x_2x_2}f_{x_2}
\end{pmatrix}.
\end{align}
The Hessian matrix of $g$, denoted by $H_g$, is given by 
\begin{equation}
   H_g=
\begin{pmatrix}
    g_{x_1x_1} & g_{x_1x_2} \\ g_{x_1x_2} & g_{x_2x_2}
\end{pmatrix},
\end{equation}
where 
\begin{align}
\label{eq:h-rv-gradient}
        &g_{x_1x_1} = 2(f_{x_1x_1}^2+f_{x_1x_1x_1}f_{x_1}+f_{x_1x_1x_2}f_{x_2}+f_{x_1x_2}^2), \nonumber \\
        &g_{x_1x_2} = 2(f_{x_1x_1x_2}f_{x_1}+f_{x_1x_1}f_{x_1x_2}+f_{x_1x_2x_2}f_{x_2}+f_{x_1x_2}f_{x_2x_2}), \nonumber \\
        &g_{x_2x_2} = 2(f_{x_1x_2x_2}f_{x_1}+f_{x_1x_2}^2+f_{x_2x_2}^2+f_{x_2x_2x_2}f_{x_2}).
\end{align}
\noindent We can express the condition for the ridge-valley graph as:
\begin{align}
\label{eq:rv-h}
    \Tilde{h} &=f_{x_1}g_{x_2}-f_{x_2}g_{x_1} \nonumber \\
    &=2(f_{x_1}(f_{x_1x_2}f_{x_1}+f_{x_2x_2}f_{x_2})-f_{x_2}(f_{x_1x_1}f_{x_1}+f_{x_1x_2}f_{x_2})) \nonumber \\
    &=2((f_{x_1}^2-f^2_{x_2})f_{x_1x_2}+f_{x_1}f_{x_2}(f_{x_2x_2}-f_{x_1x_1})) = 0.
\end{align}
Similar to the Jacobi set extraction, since $\Tilde{h}$ is derived from $f$ and its derivatives, the span setting remains the same. We can thus apply the contour extraction method by directly substituting $f$ and its derivatives with $\Tilde{h}$ and its derivatives to extract the ridge-valley graph. In the final step, according to \cite{gregory2013ridge}, instead of computing the critical points of $\Tilde{h}$ where $\Tilde{h}=0$, we compute the critical points of $f$ and incorporate them into the result. This approach ensures that all critical points of $f$ are included in the ridge-valley graph and it is more efficient than computing critical points of $\Tilde{h}$ directly. The gradient $\nabla \Tilde{h}$ can be obtained by substituting ~\cref{eq:h-rv-gradient} into ~\cref{eq:h-gradient}.

Following~\cref{sec:ridge-valley-graph}, we can classify whether a point on the ridge-valley graph corresponds to a (pseudo-)ridge or a (pseudo-)valley by checking the signs of the 2nd-order derivatives of $f$ and $g$ in the direction tangent to the level set of $f$. The direction $\mathbf{m}$ tangent to a contour of $f$ is given by \cref{eq:direction}. 
The 2nd-order derivatives of $f$ and $g$ in the $\mathbf{m}$ direction are  $f_{\mathbf{m}\mathbf{m}}=\mathbf{m}^{\top}H\mathbf{m}$ and 
$g_{\mathbf{m}\mathbf{m}}=\mathbf{m}^{\top}H_g\mathbf{m}$ respectively.~\cref{tab:classification} summarizes the classification based on the signs of $f_{\mathbf{m}\mathbf{m}}$ and $g_{\mathbf{m}\mathbf{m}}$. 
\begin{table}[!ht]
\centering
\vspace{-1mm}
\caption{Ridge-valley graph point classification based on the signs of $f_{\mathbf{m}\mathbf{m}}$ and $g_{\mathbf{m}\mathbf{m}}$.}
\begin{tabu}{*{3}{c}}
\toprule
Classification & $f_{\mathbf{m}\mathbf{m}}$ & $g_{\mathbf{m}\mathbf{m}}$ \\
\midrule
Ridge point & $<0$ & $>0$\\
Valley point & $>0$ & $>0$\\
Pseudo-ridge point & $<0$ & $<0$ \\
Pseudo-valley point & $>0$ & $<0$ \\
\bottomrule
\end{tabu}
\label{tab:classification}
\vspace{-4mm}
\end{table}

\subsection{Time Complexity}
\label{sec:complexity}

We work with 2D MFA models with degree-4 polynomials ($p=4$).
Let $n$ denote the number of spans. 
Querying the function value and the gradient at a given point takes $\mathcal{O}(p^2)$. 
Using gradient descent to locate the starting points for particle tracing takes $\mathcal{O}(c_{max}p^2)$, where $c_{max}$ is the predefined maximum iteration number of gradient descent. 
In each span, as we set $(p+3)^2$ initial points for gradient descent, finding all starting points takes $\mathcal{O}(c_{max}p^4)$. Each application of RK4 computes four gradients and takes $\mathcal{O}(p^2)$. Assume we trace $k$ points in a span, particle tracing takes $\mathcal{O}(kp^2)$. The overall time complexity for particle tracing in a span is $\mathcal{O}(kp^2+c_{max}p^4)$.
The time complexity to locate all critical points is $\mathcal{O}(i_{max}p^{4}n)$, where  $i_{max}$ is the maximum number of iterations using the Newton’s method~\cite{ma2024critical}. In a single span, there are at most $(p+3)^2$ trajectories; and establishing the connection among different trajectories and critical points takes $\mathcal{O}(p^4)$ per span.
In summary, the overall time complexity for contour extraction is 
$\mathcal{O}(nkp^2+n(c_{max}+i_{max})p^4)$. As Jacobi set and ridge-valley graph extraction are based on contour extraction from derived functions, their time complexity stays the same.

\section{Experimental Results}
\label{sec-experimental-results}

To evaluate our framework, we perform a number of experiments with MFA models representing synthetic and scientific datasets; see the supplement for details on these MFA models. 
We address the following questions: Given an MFA model as input, is it possible to extract complex topological descriptors without discretizing the entire domain? Furthermore, how does our method perform in comparison to discrete methods that rely on model discretization? 

For each MFA model, we extract contours, Jacobi sets, or ridge-valley graphs from the model, and compare the results against certain discrete methods.  

\para{Implementation.} 
The code is compiled using g++ version 11.4.0 with the -O3 optimization flag. Multithreading is achieved through Threading Building Blocks (TBB), which dynamically balances the workload across threads. The number of threads is configured to match the number of hardware cores. All our experiments are conducted on a desktop powered by an Intel i9 CPU (3.5GHz) with 8 hardware cores and 8 threads, paired with 32 GB of DDR4 RAM. 

A naive strategy for extracting topology from a continuous model is to sample it into a discrete representation and then apply established discrete methods.
For comparative analysis, we discretize the MFA model onto a mesh to obtain a discrete representation. After triangulation, we use ParaView~\cite{james2005paraView} to extract contours and the Topology ToolKit (TTK)~\cite{tierny2018topology} to extract Jacobi sets and ridge-valley graphs from the resulting discrete representation. ParaView and TTK work with piecewise-linear (PL) interpolations of scalar functions defined over triangulated meshes. In particular, we use \verb|Contour| and \verb|TTKJacobiSet| filters to extract contours and Jacobi sets from discrete representations, respectively.
The \verb|Contour| filter in ParaView implements the marching squares algorithm~\cite{lorensen1987marching} to extract the contour. 
We note that \verb|TTKJacobiSet| typically includes the entire domain boundary in the resulting Jacobi set, even though these boundary elements may not genuinely belong to it; hence, we exclude domain boundaries from the Jacobi set comparisons.
We use \verb|TTKJacobiSet| to extract the ridge-valley graph as a special type of Jacobi set. 
We refer to our approach as the \emph{MFA continuous method}, and refer to the methods based on discrete representations collectively as the \emph{discrete method}.

\begin{figure}[!ht]
\centering
\vspace{-3mm}
\includegraphics[width=0.85\linewidth]{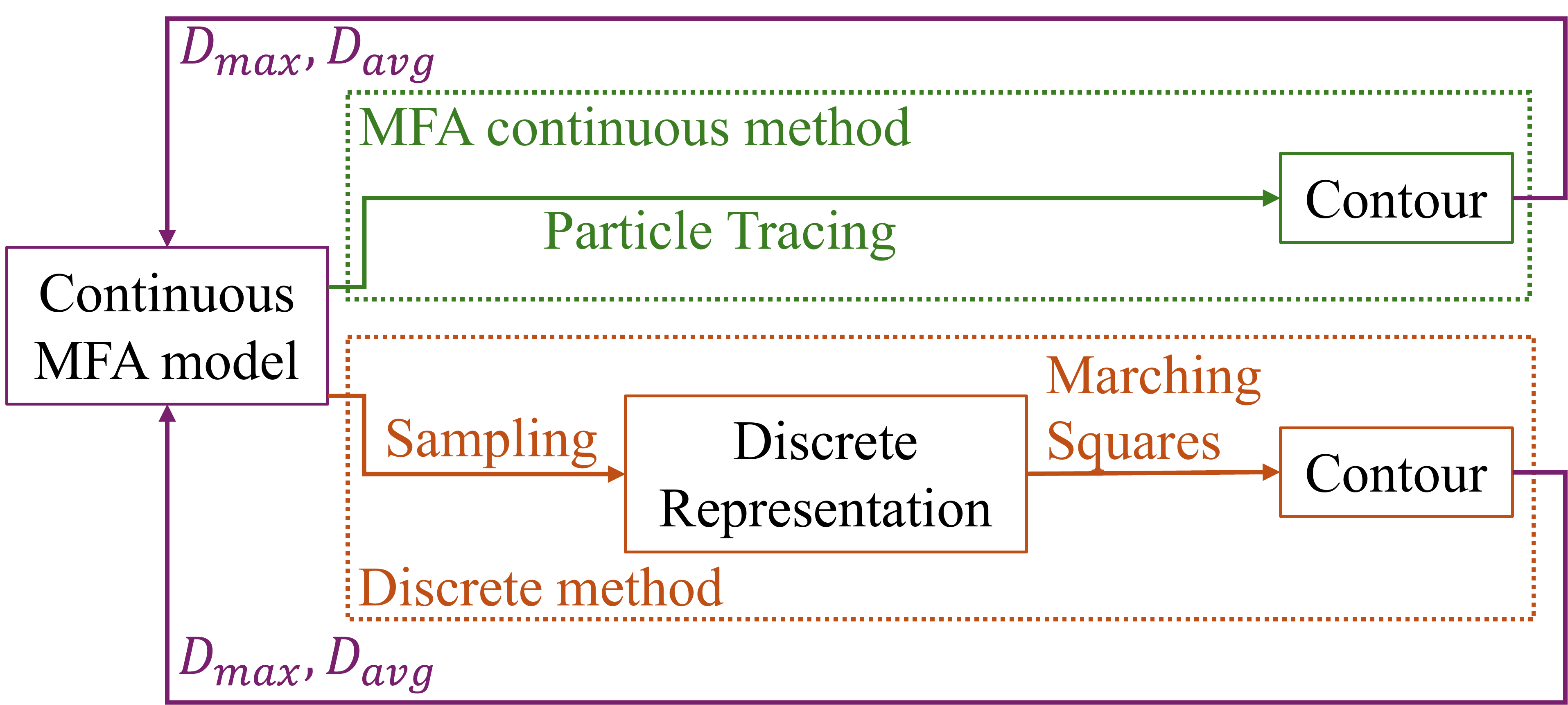}
\vspace{-3mm}
\caption{Overview of the MFA continuous method vs. the discrete method.}
\label{fig:method-flow-chart}
\vspace{-2mm}
\end{figure}

\para{Experimental setup.}
The goal of our experiments is to assess whether our MFA continuous method can effectively extract contours, Jacobi sets, and ridge-valley graphs from a continuous model, especially considering that the ground truth for these topological descriptors is typically unavailable. We compare our results to those obtained using the discrete method, which processes a discrete sample drawn from the same MFA model. The outputs from these discrete methods serve as a reference, not the ground truth, because these discrete methods operate on a PL approximation generated from a discretization of the MFA model, rather than on the MFA model itself. Despite these methodological differences, these discrete methods provide a meaningful benchmark for assessing how closely our results align with an established discrete pipeline.~\cref{fig:method-flow-chart} illustrates an overview of contour extraction using both methods.  

Our experimental procedure is summarized as follows: 
\begin{enumerate}[noitemsep,leftmargin=*]
\item Given an MFA model as the input, we apply our MFA continuous method to extract topological descriptors---contours, Jacobi sets, and ridge-valley graphs---from the model. 
\item For comparative analysis, we generate a discrete representation by sampling points from the MFA model and apply ParaView or TTK to their PL interpolation to extract the corresponding topological descriptors. 
\item We evaluate the results using metrics detailed below. 
\end{enumerate}

\para{Evaluation metrics.}
Since both approaches are derived from the same MFA model, we can assess the quality of the contours extracted by each method by measuring the difference between the MFA function value \( f(\mathbf{x}) \) and the prescribed isovalue \( a \) at each point \( \mathbf{x} \) in the domain,  
\begin{equation}
\label{eq:discrepancy}
    e(\mathbf{x}) = |f(\mathbf{x})-a|. 
\end{equation}

For Jacobi set extraction, we reformulate the task into a $0$-isovalue contour extraction of function $h$ as shown in~\cref{eq:jaocbi-set-h}, resulting in the difference measure $e(x) = |h(\bf{x})|$. Similarly, for ridge-valley graph extraction, the discrepancy is computed as $e(\bf{x}) = |\tilde{h}(\bf{x})|$, where $\tilde{h}(\bf{x})$ is given by \cref{eq:rv-h}.
We report both the maximum difference $\emax$ and average difference $\eavg$ across all points $x$. To quantitatively evaluate the extracted topological descriptors, we report the number of loops (\#Loop) and the number of connected components (\#CC). 

\subsection{An Overview of MFA Models}
\label{sec:datasets-overview}

\begin{enumerate}[noitemsep,leftmargin=*]
\item Contour: Applied to \textbf{Sinc}, \textbf{S3D}, and Schwefel.
\item Jacobi set: Applied to \textbf{Gaussian pair}, \textbf{von K\'arm\'an vortex street}, \textbf{Hurricane Isabel}, and Boussinesq approximation.
\item Ridge-valley graph: Applied to \textbf{Gaussian mixture} and \textbf{CESM}.
\end{enumerate}
Here, experiments in bold are described in this section and others are included in the supplement.

We explore four synthetic MFA models (Schwefel, Sinc, Gaussian pair, and Gaussian mixture) and five scientific MFA models (S3D, von K\'arm\'an vortex street, Hurricane Isabel, Boussinesq approximation, and CESM); see the supplement for details. 
We use synthetic MFA models to validate the accuracy of our results, and scientific MFA models to demonstrate the efficacy of our framework. The computation time is listed in \cref{tab:time}.  With 8 threads, our method achieves a speedup of approximately 7$\times$ to 8$\times$, highlighting its parallel efficiency. Although our method is slower than the discrete approach, it avoids model discretization and enables direct analysis on the continuous model. This represents a fundamentally different strategy that offers improved fidelity for downstream topological analysis.

\begin{table}[!ht]
\vspace{-3mm}
\caption{Computation time in seconds.}
\scriptsize
\centering
\setlength{\tabcolsep}{2pt} 
\begin{tabu}{c|cc|c}
\toprule
MFA Models & \multicolumn{2}{c|}{\CellWithForceBreak{MFA \\Continuous Method}} &  Discrete Method\\
\midrule 
 & Single Thread & 8 Threads & Single Thread \\ 
 \midrule
\multicolumn{4}{c}{Step Size $s=l/4$, Sampling Ratio 4}\\
\midrule
Schwefel (Contour, $a=100$) & 8.97 & 1.24  & 0.0466 \\
Sinc (Contour, $a=0.33$)& 1.49 & 0.201  &  0.0411 \\
Gaussian Pair (Jacobi set) & 1.61 & 0.233 & 0.0530 \\
\CellWithForceBreak{Gaussian Mixture \\(Ridge-valley graph)} & 50.6 & 6.43 & 0.310 \\
\midrule
\multicolumn{4}{c}{Step Size $s=l/16$, Sampling Ratio 16}\\
\midrule
S3D (Contour, $a=50$) & 17.7 & 2.37 & 5.71 \\
\midrule
\multicolumn{4}{c}{Step Size $s=l/32$, Sampling Ratio 32}\\
\midrule
K\'arm\'an (Jacobi set) & 131 & 17.7 & 10.9 \\
Boussinesq (Jacobi set) & 557 & 76.6 & 23.2 \\
CESM (Ridge-valley graph) & 3218 & 408 & 115 \\
\midrule
\multicolumn{4}{c}{Step Size $s=l/64$, Sampling Ratio 64}\\
\midrule
Hurricane (Jacobi set) & 6289 & 857 & 394 \\
\bottomrule
\end{tabu}
\label{tab:time}
\vspace{-3mm}
\end{table}

\subsection{Sinc: Contour Extraction}
\label{sec:sinc-contour}

\para{Parameter selection via ablation studies.}~We select our parameters---step size $s$, accuracy threshold $\epsilon$, and trajectory connection threshold $\gamma$---based on a series of ablation studies. Recall from~\cref{sec:MFA-concept} that the \emph{span length} $l$ is the distance in the MFA model between knots and corresponding control points. All the MFA models in this study have a uniform span length across the entire model. 

\begin{table}[!ht]
\vspace{-3mm}
\caption{Sinc model: evaluation of contour extraction with various isovalues ($a$). GT denotes the ground truth.}
\scriptsize
\centering
\begin{tabu}{c|cccccc}
\toprule
$a$ & $\emax$ & $\eavg$ & \#Loop & GT\#Loop  & \#CC  & GT\#CC\\ 
\midrule
0.33 & $9.8e^{-11}$ & $5.5e^{-12}$ & 40 & 40 & 60 & 60 \\
0.79 & $9.9e^{-11}$ & $5.1e^{-12}$ & 28 & 28 & 32 & 32\\
\bottomrule
\end{tabu}
\label{tab:theoritical-contour}
\vspace{-3mm}
\end{table}

For all synthetic and scientific MFA models, we use a model-specific step size in each experiment, selected based on an ablation study. We decrease step size $s \in \{l/2, l/4, l/8, \dots, l/2^{p}, \dots\}$ until both \#Loop and \#CC reach convergence. 
The accuracy threshold is chosen at $\epsilon = 1e^{-10}$ for all experiments. Trajectories are connected when the distance between them is within a connection threshold $\gamma = 2s$ as described in \cref{sec:connecting-step}; see the supplement for further details on parameter selection.

\begin{figure}[!ht]
\centering
\includegraphics[width=0.8\linewidth]{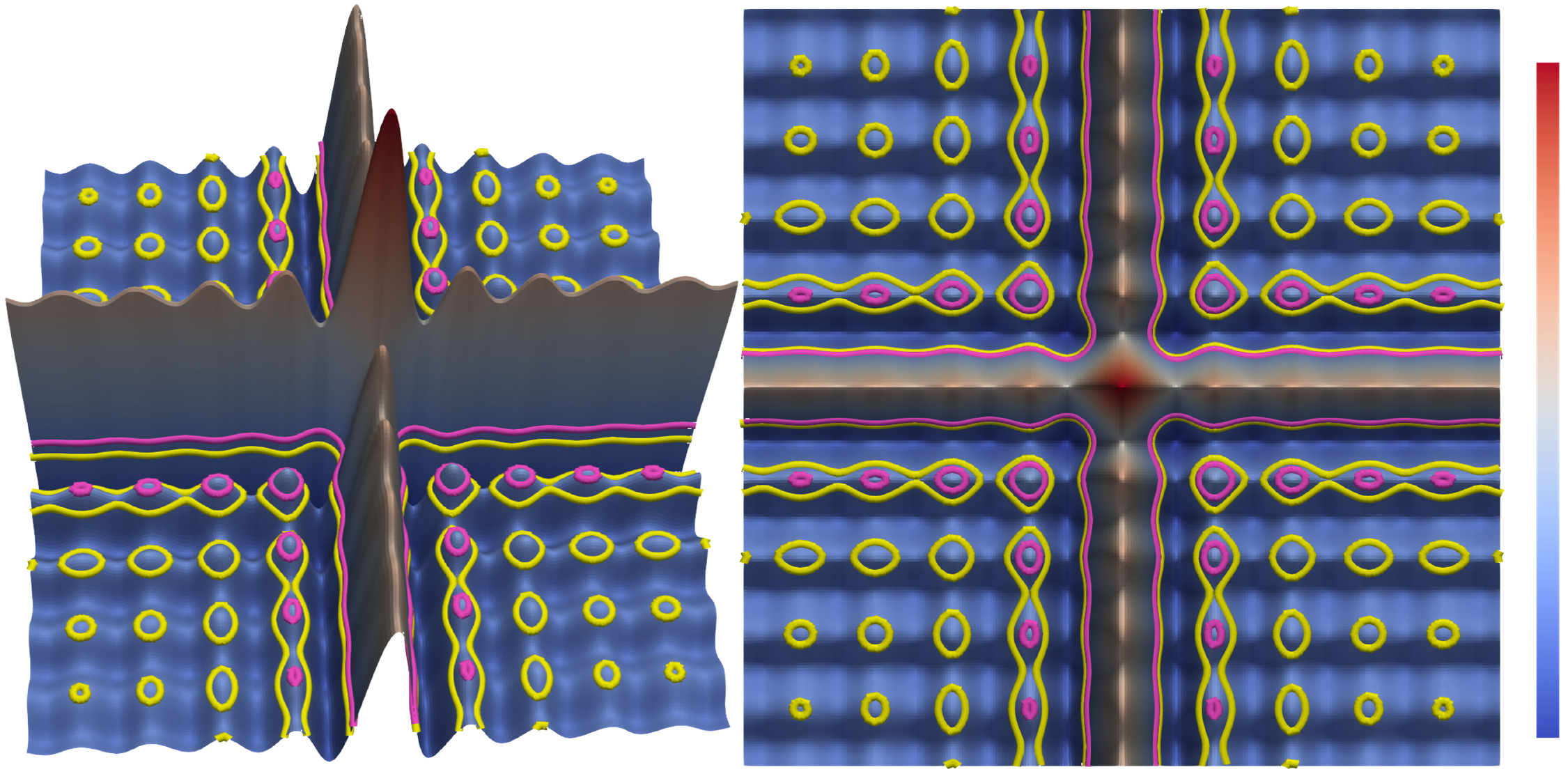}
\vspace{-4mm}
\caption{Sinc model: contour extraction with isovalues $a = 0.33$ (yellow) and $a = 0.79$ (pink). Results are shown in different views. }
\vspace{-1mm}
\label{fig:sinc-contour}
\end{figure}

\begin{figure}[!ht]
\centering
\includegraphics[width=\linewidth]{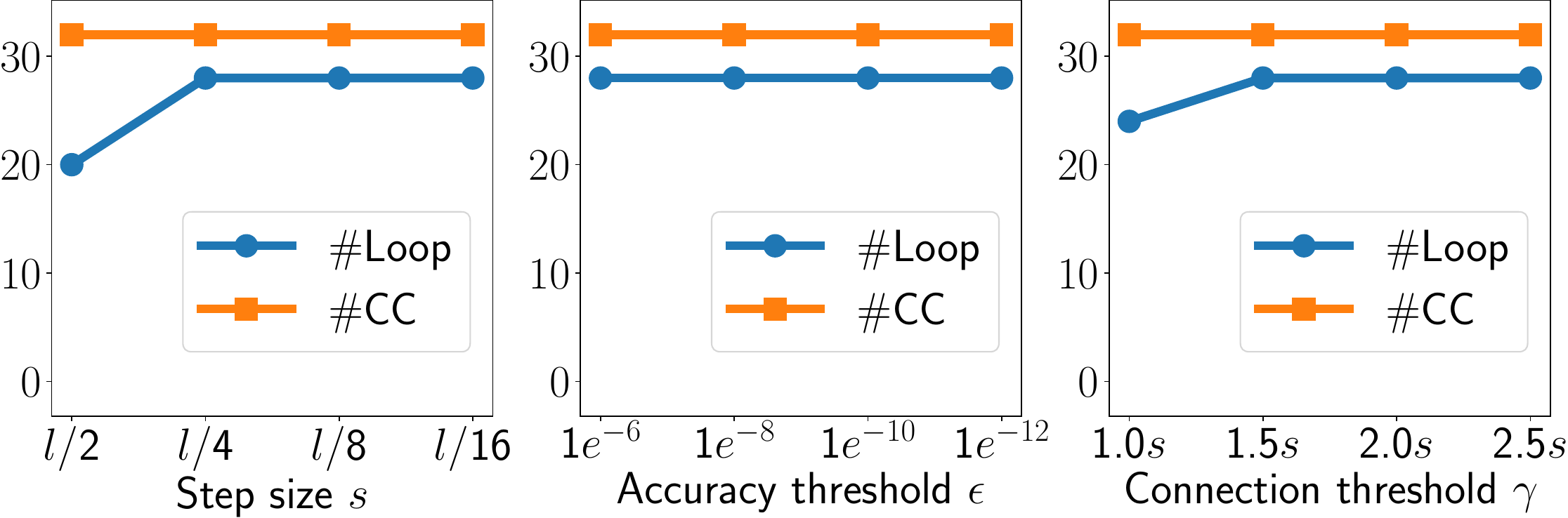}
\vspace{-7mm}
\caption{Contour extraction of Sinc model at $a=0.79$. Number of loops and connected components vs. step size $s$ (left), accuracy threshold $\epsilon$ (middle), and trajectory connection threshold $\gamma$ (right).}
\vspace{-6mm}
\label{fig:sinc-parameters}
\end{figure}

When constructing a discrete representation from an MFA model, we set the grid spacing equal to the chosen step size $s$. This ensures that the edge lengths of the PL contour in the discrete representation remains consistent with those extracted from the continuous representation. 
For example, a grid spacing of \(l/4\) results in sampling 4 points per dimension within each span, corresponding to a sampling ratio of 4.

\begin{figure*}[!ht]
\centering
\includegraphics[width=\linewidth]{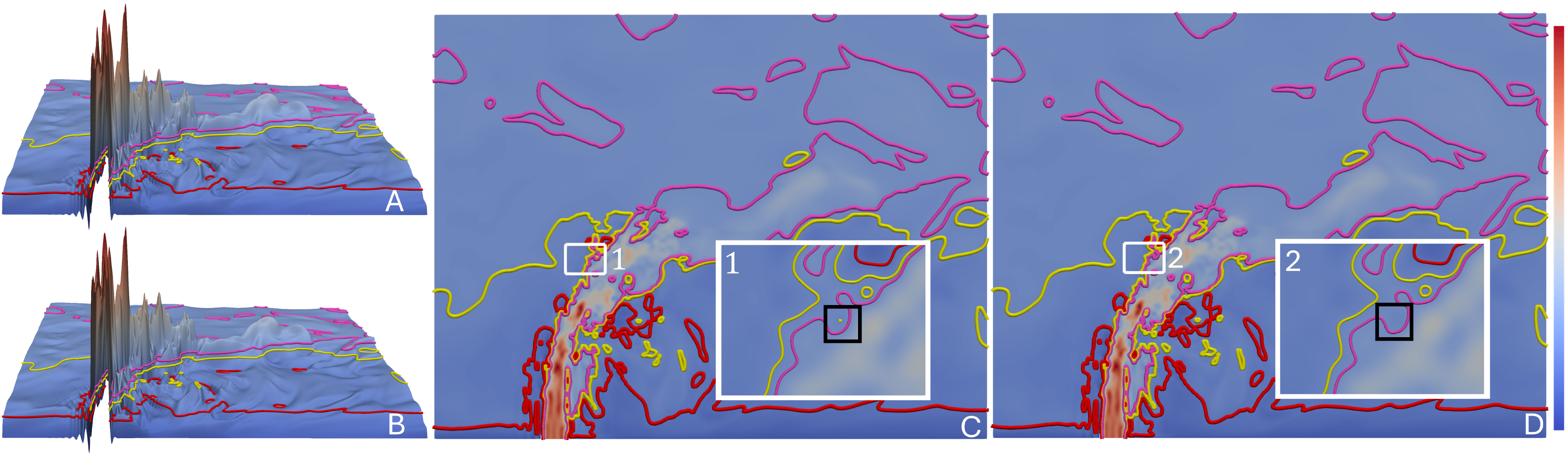}
\vspace{-9mm}
\caption{S3D model: contour extraction using isovalues $a=30$ (red), $a=50$ (yellow), and $a=60$ (pink). (A, C): results from MFA continuous method with a step size of $s=l/16$, viewed from the top and the side; (B, D): results from discrete method with a sampling ratio of 16, viewed from the top and the side. (1) and (2) provide the zoomed-in views of the white blocks in (C) and (D), respectively.}
\label{fig:s3d-contour}
\vspace{-2mm}
\end{figure*}

In \cref{fig:sinc-contour}, contours are extracted from the Sinc model using isovalues $a=0.33$ and $a=0.79$. The original Sinc function serves as the ground truth. \cref{tab:theoritical-contour} shows that the errors at all nodes are below the threshold $\epsilon = 1e^{-10}$. The numbers of loops and connected components align exactly with those of the ground truth. In \cref{fig:sinc-parameters}, we illustrate the results for varying step sizes ($s$), thresholds ($\epsilon$), and connection thresholds ($\gamma$) at $a=0.79$.
For the Sinc model, step size $s=l/4$ is sufficient to ensure convergence for the number of connected components and loops.
Under these conditions, the results demonstrate robustness with respect to variations in $\epsilon$ and $\gamma$.

\subsection{S3D: Contour Extraction}
\label{sec:s3d-contour}
\cref{tab:s3d-contour-1} presents an evaluation of our MFA continuous method with a step size of $s=l/16$ alongside the discrete method with a sampling ratio of 16, using the original continuous MFA model as a reference. The discrete method differs primarily due to discretization artifacts. At $a=50$, it misses a small loop that is visible in the black block of \cref{fig:s3d-contour} (1) but absent in \cref{fig:s3d-contour} (2).
In \cref{fig:s3d-contour}, the three contours at isovalues $a=30, 50, 60$ from the S3D model are displayed using both continuous and discrete representations.

\begin{figure}[!th]
\centering
\includegraphics[width=\linewidth]{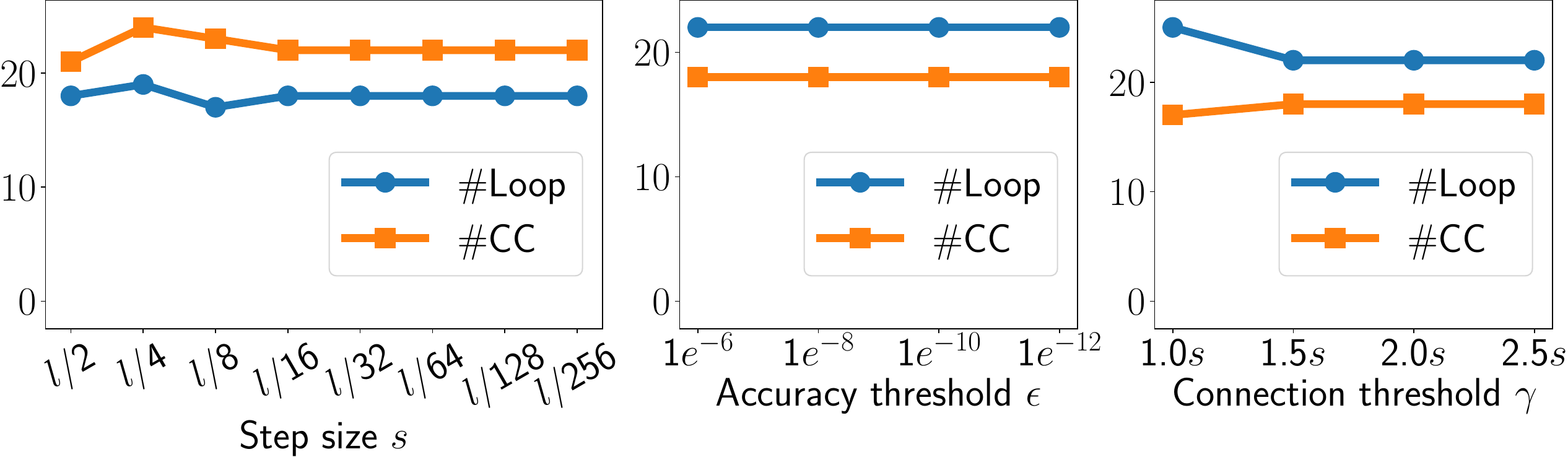}
\vspace{-6mm}
\caption{Contour extraction of S3D model at $a=50$. Number of loops and connected components vs. step size $s$ (left), accuracy threshold $\epsilon$ (middle), and connection threshold $\gamma$ (right).}
\label{fig:s3d-contour-parameters}
\end{figure}

In \cref{fig:s3d-contour-parameters}, results are visualized for different step sizes ($s$), thresholds ($\epsilon$), and connection thresholds ($\gamma$) at $a=50$. A step size of $s=l/16$ is chosen to guarantee convergence of both connected components and loops. This selection yields robust results with respect to variations in $\epsilon$. Additionally, $\gamma=2s$ is adopted for consistency across experiments and to ensure convergence. 

\begin{table}[!ht]
\caption{S3D model: contour extraction with step size $s=l/16$ and a sampling ratio 16.}
\scriptsize
\centering
\setlength{\tabcolsep}{3pt} 
\begin{tabu}{c|cccc|cccc}
\toprule
\ & \multicolumn{4}{c|}{MFA Continuous Method} &  \multicolumn{4}{c}{Discrete Method}\\
\midrule 
$a$ & $\emax$ & $\eavg$  & \#Loop & \#CC & $\emax$  & $\eavg$  & \#Loop & \#CC \\ 
30 & $1.0e^{-10}$& $5.6e^{-12}$ & 21 & 23 & 0.25 & $8.5e^{-3}$ & 21 & 23 \\
50 & $1.0e^{-10}$ & $4.5e^{-12}$ &18 & 22 & 0.23 & $1.1e^{-2}$ & 17 & 21\\
60 & $1.0e^{-10}$& $1.0e^{-11}$ & 13& 21 & 0.22 & $6.4e^{-3}$ & 13 & 21 \\
\bottomrule
\end{tabu}
\label{tab:s3d-contour-1}
\vspace{-3mm}
\end{table}

\begin{table}[!ht]
\caption{Jacobi set extraction of MFA models.} 
\scriptsize
\centering
\setlength{\tabcolsep}{3pt} 
\begin{tabu}{cccc|cccc}
\toprule
\multicolumn{4}{c|}{MFA Continuous Method} &  \multicolumn{4}{c}{Discrete Method}\\
\midrule 
$\emax$ & $\eavg$ &\#Loop & \#CC & $\emax$ & $\eavg$ &\#Loop & \#CC \\ 
\midrule
\multicolumn{8}{c}{Gaussian Pair, Step Size $s=l/4$, Sampling Ratio 4}\\
\midrule
$6.8e^{-11}$& $4.3e^{-12}$ & 0 & 2 & 0.31 & $2.2e^{-2}$ & 161 & 2 \\
\midrule
\multicolumn{8}{c}{K\'arm\'an, Step Size $s=l/32$, Sampling Ratio 32}\\
\midrule
$1.0e^{-10}$& $2.2e^{-11}$ & 82 & 124 & 1.3 & $1.5e^{-2}$ & 5039  & 559 \\
\midrule
\multicolumn{8}{c}{Hurricane, Step Size $s=l/64$, Sampling Ratio 64}\\
\midrule
$1.0e^{-10}$& $2.9e^{-11}$ & 1901 & 2015 & 1.0 & $4.2e^{-3}$ & $4.1e^5$ & 1956 \\
\bottomrule
\end{tabu}
\label{tab:jacobi-set}
\vspace{-6mm}
\end{table}

\subsection{Gaussian Pair: Jacobi Set Extraction}
Given a pair of scalar functions $f$ and $g$ represented by MFA models, \cref{fig:teaser-image} presents the Jacobi sets $\mathbb{J}:=\mathbb{J}(f,g)$ extracted from both the continuous Gaussian pair model (C) and its discrete representation (F). The scalar functions $f$ and $g$ are depicted as purple and red contours, respectively. 
Results from discrete representations exhibit visible artifacts due to discretization: most notably, they exhibit numerous zigzag patterns, in comparison with significantly smoother results from the continuous MFA model. 

Quantitative evaluation is presented in~\cref{tab:jacobi-set}. 
Due to artifacts arising from discretizing a continuous model, the discrete method exhibits significantly larger error compared to the MFA continuous method. Specifically, zig-zag patterns result in numerous small triangles, causing the discrete method to generate many spurious loops.

\begin{figure}[!th]
\centering
\vspace{-1mm}
\includegraphics[width=\linewidth]{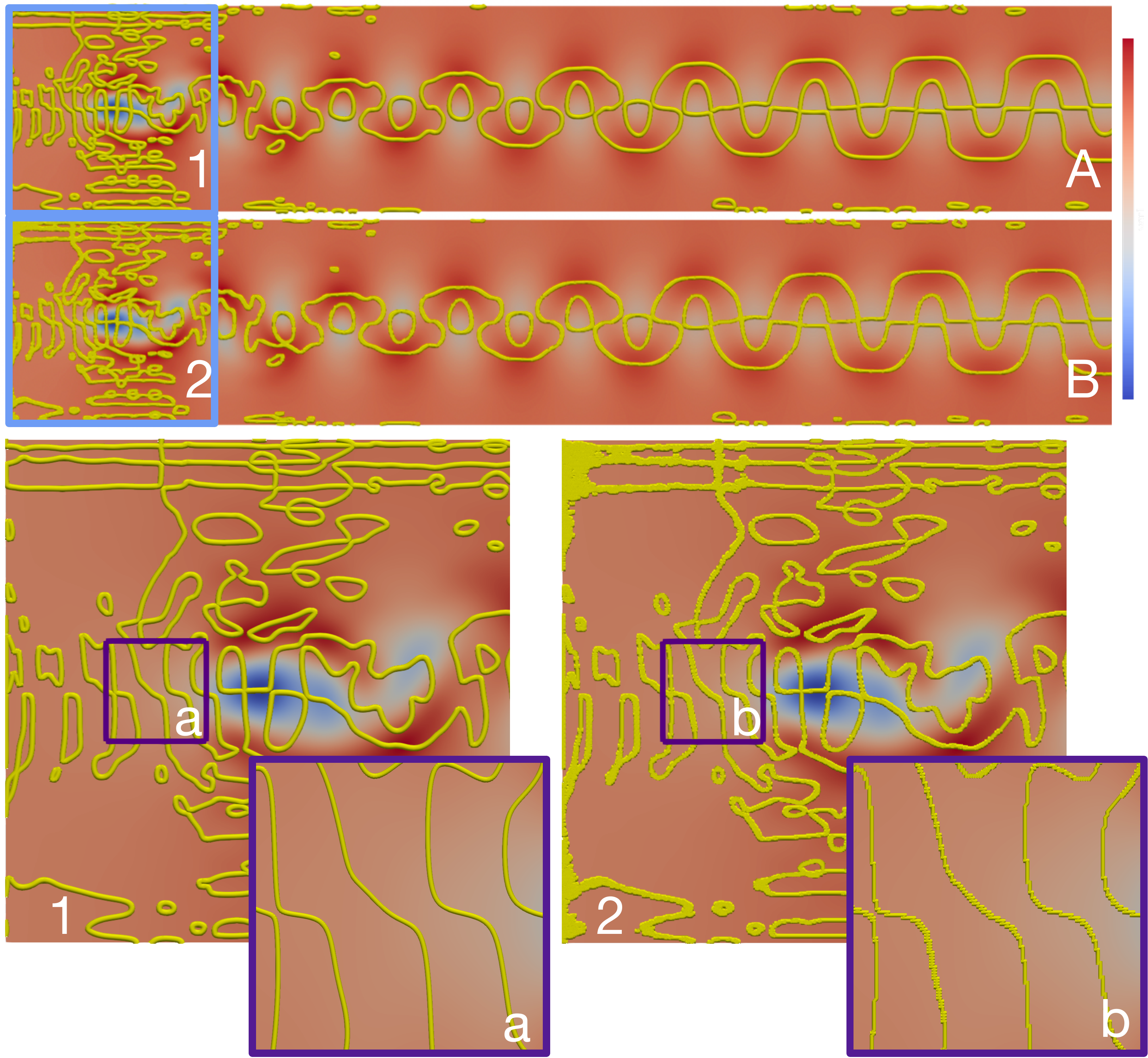}
\vspace{-7mm}
\caption{Von K\'arm\'an vortex street model: Jacobi set extraction using MFA continuous method (A) and discrete method (B).  
(1) and (2): zoomed-in views of blue blocks in (A) and (B), respectively. (a) and (b): zoomed-in views of purple blocks in (1) and (2), respectively.}
\vspace{-4mm}
\label{fig:Karman-vortex-street}
\end{figure}

\subsection{Von K\'arm\'an Vortex Street: Jacobi Set Extraction}

\cref{fig:Karman-vortex-street} presents a visualization of the von K\'arm\'an vortex street MFA model at time step 1500. The step size is $s=l/32$, and the sampling ratio is $32$. 
Due to resolution limitations, the discrete representation (B) exhibits pronounced zigzag patterns, while the continuous representation (A) is significantly smoother. These differences are highlighted in zoomed-in views of (1) and (2), where zigzag patterns from the discrete representation are clearly visible in block (b) of (2), c.f.,~block (a) of (1). 
In~\cref{tab:jacobi-set}, the error of the MFA continuous method remains below $\epsilon=1e^{-10}$, whereas the error from the discrete method is much larger. 
The discrete representation generates many spurious loops due to these zig-zag patterns.

\begin{figure}[!ht]
\centering
\includegraphics[width=1.0\linewidth]{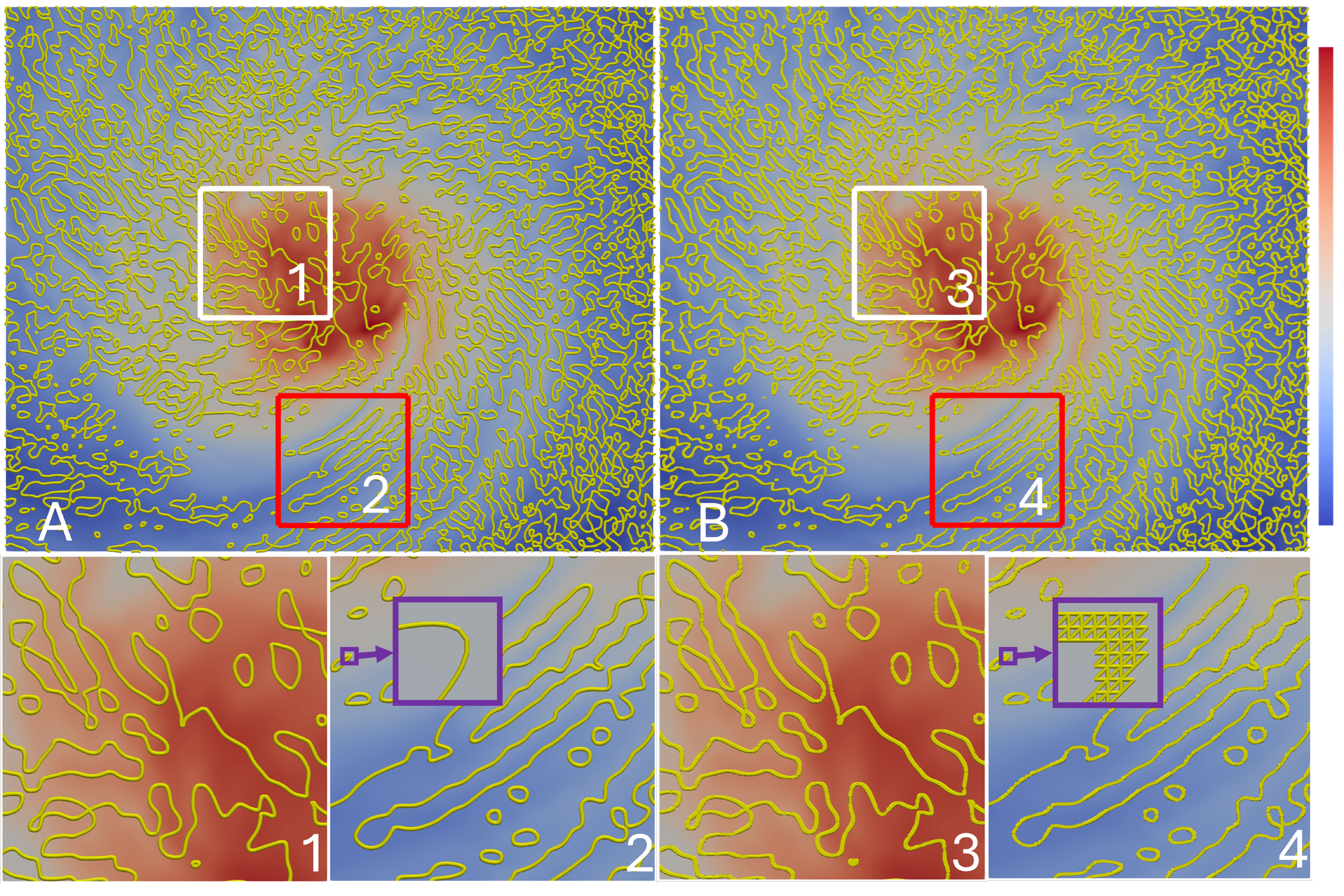}
\vspace{-6mm}
\caption{Hurricane Isabel model: Jacobi set extraction using MFA continuous method (A) and discrete method (B). (1-4): zoomed-in views of the white (1,3) and red (2,4) blocks, respectively. Purple insets further magnify the highlighted regions indicated by arrows.}
\vspace{-2mm}
\label{fig:hurricane}
\end{figure}

\subsection{Hurricane Isabel: Jacobi Set Extraction}

In \cref{fig:hurricane}, we present the Jacobi set extracted from the Hurricane Isabel model computed between temperature and pressure fields. Results are shown for the region $[125,375]\times[120,330]$. We select a slice of the model at a height of 50, where the hurricane exhibits significant spatial expansion and displays many characteristic structures~\cite{klotzl2022local}. 
The step size for MFA continuous method is $s=l/64$, and the corresponding sampling ratio is $64$.
The results from the continuous representation in blocks (1) and (2) are smoother than the results from the discrete representations in blocks (3) and (4), respectively. 
Zig-zag patterns, like those in the purple block of (4), contribute to the high number of loops in the discrete representation, as shown in \cref{tab:jacobi-set}.

\begin{table}[!ht]
\caption{Ridge-valley graph extraction from MFA models.}
\scriptsize
\centering
\setlength{\tabcolsep}{3pt} 
\begin{tabu}{cccc|cccc}
\toprule
\multicolumn{4}{c|}{MFA Continuous Method} &  \multicolumn{4}{c}{Discrete Method}\\
\midrule
$\emax$ & $\eavg$ &\#Loop & \#CC & $\emax$ & $\eavg$ &\#Loop & \#CC \\ 
\midrule
\multicolumn{8}{c}{Gaussian Mixture, Step Size $s=l/4$, Sampling Ratio 4}\\
\midrule
$9.9e^{-11}$& $5.9e^{-12}$ & 2 & 1 & $1.4$ & $4.3e^{-1}$ & 970 & 1 \\
\midrule
\multicolumn{8}{c}{CESM, Step Size $s=l/32$, Sampling Ratio 32}\\
\midrule
$1.0e^{-10}$& $1.9e^{-11}$ & 3402 & 860 & $4.2e^{2}$ & $2.7e^{2}$ & $2.1e^{5}$ & 751 \\
\bottomrule
\end{tabu}
\vspace{-4mm}
\label{tab:rv-graph}
\end{table}

\subsection{Gaussian Mixture: Ridge-Valley Graph Extraction}

\cref{fig:teaser-image} presents the results of ridge-valley graph extraction using the MFA continuous method (B) and the discrete method (E). Based on a convergence study, the step size for (B) is set to \( s = l/4 \), and the sampling ratio for (E) is 4. While the overall ridge-valley graph structures align for both methods, the discrete representation in (E) exhibits pronounced zigzag patterns and generates hundreds of spurious loops (see \#Loop in~\cref{tab:rv-graph}). Moreover, the discrete method shows a significantly larger error compared to the original MFA model (see \( \emax \) and \( \eavg \) in~\cref{tab:rv-graph}).

\begin{figure}[!ht]
\centering
\vspace{-3mm}
\includegraphics[width=\linewidth]{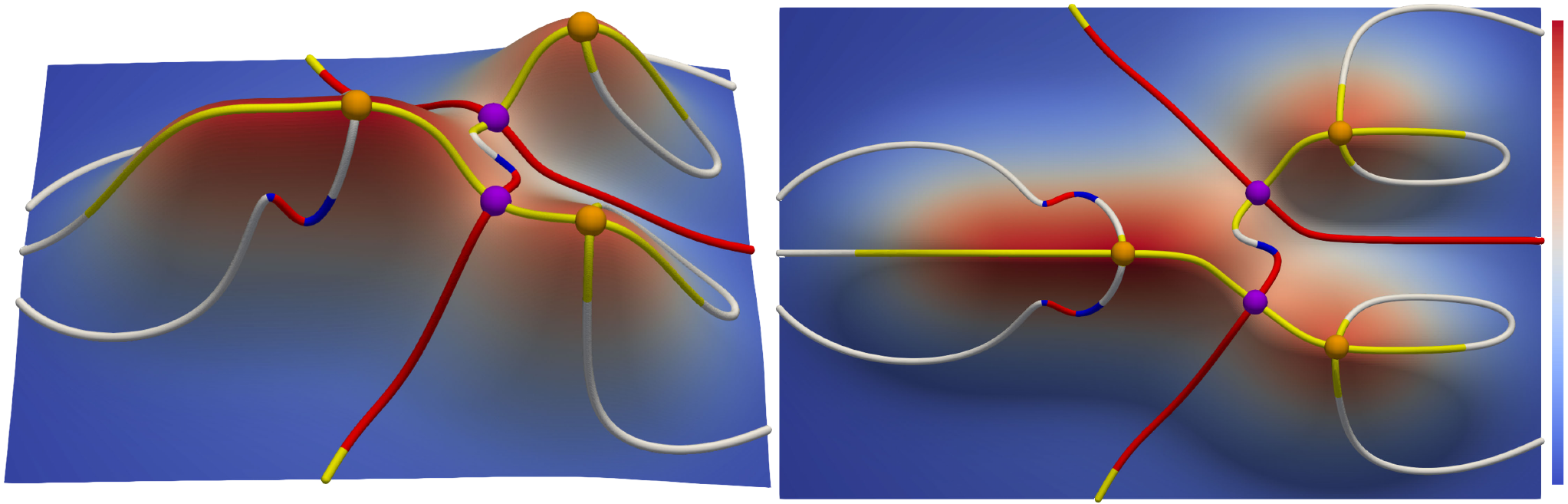}
\vspace{-6mm}
\caption{Gaussian mixture model: Ridge-valley graph extraction and classification with 3D (left) and 2D (right) views. Local maxima (orange) and saddles (purple) are connected by ridges (yellow), valleys (red), pseudo-ridges (white), or pseudo-valleys (blue).}
\vspace{-3mm}
\label{fig:gaussian-mixture-classification}
\end{figure}

By utilizing the advantages of a continuous representation that allows access to second-order derivatives, our method enables the classification of any point on the ridge-valley graph based on its position; see~\cref{tab:classification}. In~\cref{fig:gaussian-mixture-classification}, we present the classification of arcs from the extracted ridge-valley graph.

\begin{figure}[!ht]
\centering
\includegraphics[width=\linewidth]{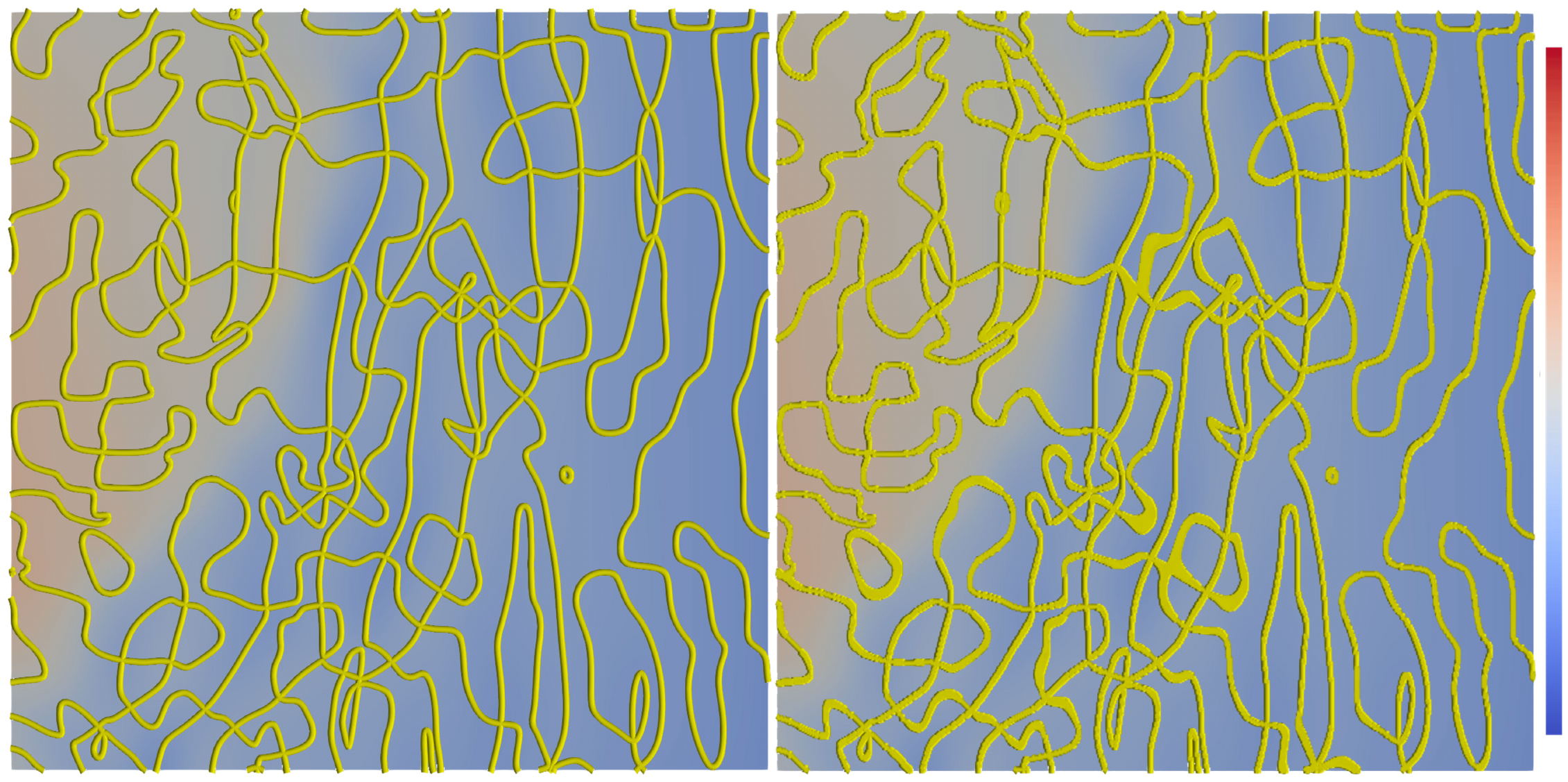}
\vspace{-6mm}
\caption{CESM model: Ridge-valley graph extraction for block $[2087,2411]\times[1367,1691]$ using MFA continuous method (left) and discrete method (right).}
\vspace{-3mm}
\label{fig:cesm}
\end{figure}

\begin{figure}[!ht]
\centering
\vspace{-1mm}
\includegraphics[width=\linewidth]{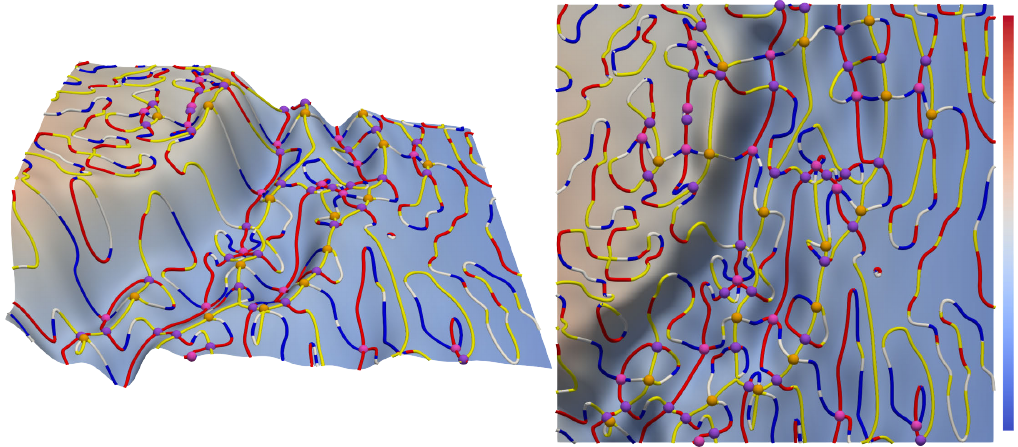}
\vspace{-6mm}
\caption{CESM model: Ridge-valley graph extraction and classification with 3D (left) and 2D (right) views. Local maxima (orange), saddles (purple), and local minima (pink) are connected by ridges (yellow), valleys (red), pseudo-ridges (white), or pseudo-valleys (blue).}
\vspace{-5mm}
\label{fig:cesm-classification}
\end{figure}

\subsection{CESM: Ridge-Valley Graph Extraction}

\cref{fig:cesm} presents the results of ridge-valley graph extraction, with a step size $s=l/32$ for the MFA continuous method and a sampling ratio of $32$ for the discrete method. Our method yields smoother results compared to the discrete representation, which introduces zigzag artifacts and numerous loops, as reported in \cref{tab:rv-graph}.
These discrepancies between the continuous model and its discretization contribute significantly to the error observed for the discrete method in \cref{tab:rv-graph}.
In \cref{fig:cesm-classification}, we present the classification of arcs from the extracted ridge-valley graph based on the MFA continuous method.

\section{Conclusion and Discussion}
\label{sec:conclusion}

We present a novel framework for extracting contours, Jacobi sets, and ridge-valley graphs directly from continuous implicit models that allow queries of function values and high-order derivatives. In our experiments, we focus on MFA models as examples of continuous implicit models. Our framework directly works on MFA models without requiring discretization, and leverages multithreading for enhanced performance. We demonstrate the effectiveness of our framework across various scientific datasets, showcasing its capability to support topological data analysis with continuous implicit models. In the future, we would like to extend our framework to extract other topological descriptors such as Morse/Morse--Smale complexes, which are similar but not quite the same as the ridge-valley graphs. Furthermore, extending these features to 3D domains is a promising and impactful direction for future research.

\para{Limitations.} 
During particle tracing, we identify multiple trajectories within each span that need to be connected. Our connecting strategy may introduce incorrect connections when trajectories are close enough, and the tracing process may produce undesirable gaps between trajectories. Reducing the step size until convergence of the number of connected components and loops will mitigate these issues, as shown by the ablation studies. 
Finally, the selection of initial points (for locating starting points during particle tracing) may affect the results. For certain MFA models, a greater number of initial points may be required within a span to ensure that starting points exist for every piece of the contours in that span. Developing a clear strategy with such guarantees is left for future work.

\acknowledgments{
This work was supported by the U.S. Department of Energy (DOE), Office of Science, Office of Advanced Scientific Computing Research, under contract numbers DE-AC02-06CH11357, DE-SC0023157, and DE-SC0022753, program manager Hal Finkel. It was also supported in party by National Science Foundation (NSF) grant DMS-2301361.}

\bibliographystyle{abbrv-doi}
\bibliography{refs-mfa}

\clearpage
\appendix 

This supplement provides details on the MFA models (\cref{sec:models}), the parameter selection (\cref{sec:parameter-selection}), and additional experimental results (\cref{sec:additional-experiments}).

\section{Details on MFA Models}
\label{sec:models}

To fit an MFA model, a set of points is given with function values observed at each point. Based on the mathematical formulations of Sec.~4, we use MFA models of degree 4 ($p=4$). 
We then tackle the following key question: given an MFA model as the input, how can we extract topological descriptors from the model, without resorting to discretization or sampling?

In \cref{tab:datasets}, we describe the MFA models representing four synthetic (left) and five scientific datasets (right); in particular, we list the number of spans used in each model.

\begin{table}[!ht]
  \caption{Details on the MFA models: the number of spans (\#Span) used in each model.}
  \scriptsize
  \centering
  \resizebox{1.0\columnwidth}{!}{
  \begin{tabu}{cc|cc}
  \toprule
  Model & \#Span & Model & \#Span\\
  \midrule
Schwefel & $71\times 71$ & S3D & $136 \times 104$ \\  
Gaussian Mixture & $46\times 71$  & Von K\'arm\'an & $156\times 16$  \\
Sinc & $27\times 27$ & Boussinesq & $46\times 146$  \\
  Gaussian Pair & $17\times 11 $ & CESM & $200 \times 100$ \\
  & & Hurricane Isabel & $162\times 162$\\
  \bottomrule
  \end{tabu}
  }
  \label{tab:datasets}
  \vspace{-2mm}
\end{table}

\para{Schwefel MFA model.}
The (scaled) Schwefel function is a non-convex benchmark function that can be defined in any dimension \cite{schwefel1981numerical}. In the 2D case, for a vector $\mathbf{x}=(x_1, x_2)^{\top}$, the function is defined as: 
\begin{equation}
\label{eq:schwefel}
    f(\mathbf{x})=\frac{1}{2}\left(418.9829 d - x_1\sin(\sqrt{|x_1|})-x_2\sin(\sqrt{|x_2|})\right).
\end{equation} 

We create the Schwefel MFA model by fitting an MFA to a 2D Schwefel function within the domain $\left[-(10.5\pi)^2, (10.5\pi)^2\right]^2$. 

\para{Gaussian Mixture MFA model.}
The function associated with a synthetic Gaussian mixture~\cite{gregory2013ridge} is defined as
\begin{align}
        f(\mathbf{x}) = & \exp\left[-8(x_1 + 0.4)^2 - 4x_2^2\right] 
 + \exp\left[-8(x_1 - 0.5)^2 - 4x_2^2\right] \nonumber \\
& + \exp\left[-8x_1^2 - 4(x_2 - 0.77)^2\right] \\
&+ \exp\left[-8x_1^2 - 4(x_2 - 1.5)^2\right] \nonumber \\
& + 0.2\exp\left[-0.3x_1^2 - 0.3(x_2 - 0.5)^2\right],
\end{align}
for $\mathbf{x}=(x_1,x_2)^{\top}$.
To generate the Gaussian mixture MFA model, we fit an MFA to the original function in the domain $[-1,1]\times[-0.8,2.3]$.

\para{Sinc MFA model.}
We define a synthetic functions of $\mathbf{x}=(x_1, x_2)^{\top}$ by adding Sinc functions in each dimension:
\begin{equation}
   f(\mathbf{x})  = \frac{sin(5x_1)}{x_1}+\frac{sin(5x_2)}{x_2}.   
\end{equation}
To generate the Sinc MFA model, we fit an MFA to $f(\mathbf{x})$ in the domain $[-2\pi,2\pi]^2$.

\para{Gaussian Pair MFA model.}
We define a pair of synthetic functions of $\mathbf{x}=(x_1, x_2)^{\top}$, one is a Gaussian function, the other is a mixture of two Gaussian functions:  
\begin{equation}
\begin{aligned}
   f(\mathbf{x}) = & 0.25 \cdot \exp\left[-\frac{(x_1 - 0.5)^2}{0.02} - \frac{(x_2 - 0.4)^2}{0.02}\right], \\
   g(\mathbf{x}) = & 0.25 \cdot \exp\left[-\frac{(x_1 -0.3)^2}{0.02} - \frac{(x_2 - 0.2)^2}{0.02}\right] \\
& + 0.25 \cdot \exp\left[-\frac{(x_1 - 0.75)^2}{0.02} - \frac{(x_2 - 0.25)^2}{0.0288}\right].
\end{aligned}  
\end{equation}
To generate the Gaussian Pair MFA model, we fit an MFA to the original functions in the domain $[0.1,0.9]\times[0.0,0.6]$. The MFA model represents $f$ and $g$ using the same spans but with different control points.

\para{S3D MFA model.}
The dataset originates from an S3D turbulent combustion simulation~\cite{chen2009terascale}, which simulates the interaction between a fuel jet combustion and an external cross-flow~\cite{fric1994vortical,grout2011direct,ma2024critical}.
We use the magnitude of the 3D velocity field as the scalar function of interest. 
We extract a 2D slice from the dataset (defined on a grid) and fit an MFA model to it, which we refer to as the S3D MFA model.

\para{Von K\'arm\'an Vortex Street MFA models.}
We work with a simulated von K\'arm\'an vortex street dataset~\cite{popinet2004free,guenther2017generic}. It originates from a simulation of a viscous 2D flow around a cylinder, capturing velocity fields over 1501 time steps. For the Jacobi set computation, we use the flow velocity magnitudes from two consecutive time steps as scalar fields $f$ and $g$ defined on a 2D grid. Specifically, we focus on time steps 1500 and 1501, where the vortex street is fully developed. We fit two MFA models to $f$ and $g$ respectively, which we refer to as the Von K\'arm\'an Vortex Street MFA models. 

\para{Boussinesq Approximation MFA model.}
The Boussinesq approximation dataset simulates a 2D flow generated by a heated cylinder \cite{popinet2004free,guenther2017generic}. It consists of 2001 time steps. For the Jacobi set computation, we use the flow velocity magnitudes from time steps 2000 and 2001 as scalar fields $f$ and $g$. We replace these scalar fields with MFA models as the surrogates.  

\para{CESM MFA model.}
The Community Earth System Model (CESM) provides extensive global climate data~\cite{neale2010description}. In our experiment, we focus on the FLDSC variable, which records the clear-sky downwelling long-wave flux on the surface, as modeled by the Community Atmosphere Model (CAM) developed at the National Center for Atmospheric Research (NCAR)~\cite{neale2010description}. We fit an MFA model to this dataset as the surrogate.  

\para{Hurricane Isabel MFA model.}
The Hurricane Isabel dataset generated using the Weather Research and Forecast (WRF) model~\cite{hurricane} provides a collection of 3D scalar fields and a velocity vector field. These fields are defined over 48 time steps. For our analysis, we focus on the scalar fields of temperature and pressure at time step 30 with height 50. As shown in~\cite{klotzl2022local}, at the height of 50, the hurricane exhibits significant spatial expansion and displays many characteristic structures. We again fit an MFA model to this dataset as the surrogate. 
\section{Parameter Selection}
\label{sec:parameter-selection}

Recall that we select our parameters---step size $s$, accuracy threshold $\epsilon$, and trajectory connection threshold $\gamma$---based on a series of ablation studies. The \emph{span length} $l$ is the distance in the MFA model between knots and corresponding control points. All the MFA models in this study have a uniform span length across the entire model. 

We first discuss the selection of step size $s$ for particle tracing and the number of initial points needed to find starting points for particle tracing.

\para{Step size.}~Particle tracing is performed independently within each span. To ensure particle tracing can be performed in every span, the step size $s$ must satisfy $s \leq \frac{l}{2}$.
To determine the proper step size, we run a series of experiments with different step sizes across each MFA model. In particular, we decrease step size $s \in \{l/2, l/4, l/8, \dots, l/2^{p}, \dots\}$ until both \#Loop (the number of loops) and \#CC (the number of connected components) reach convergence. 

For experiments involving synthetic MFA models, as shown in \cref{fig:schwefel-1-ablation-study,fig:schwefel-2-ablation-study,fig:sinc-1-ablation-study,fig:sinc-2-ablation-study,fig:gaussian-pair-ablation-study,fig:gaussian-mixture-ablation-study}, both \#Loop and \#CC converge when $s=l/4$. Therefore, we set the step size $s=l/4$. On the other hand, scientific MFA models exhibit greater variability, so we select an optimal step size individually for each model-task pair based on analogous convergence criteria. The chosen step sizes are summarized in \cref{tab:step-size}.

\begin{table}[!ht]
\caption{Step sizes for MFA model-task pairs.}
\scriptsize
\centering
\setlength{\tabcolsep}{2pt} 
\begin{tabu}{cc}
\toprule
MFA Model & Step Size $s$ \\
\midrule
Schwefel (Contour) & $l/4$ \\
Sinc (Contour) & $l/4$  \\
Gaussian Pair (Jacobi set) & $l/4$ \\
Gaussian Mixture (Ridge-valley graph) & $l/4$ \\
S3D (Contour) & $l/16$ \\
K\'arm\'an (Jacobi set) & $l/32$ \\
Boussinesq (Jacobi set) & $l/32$ \\
CESM (Ridge-valley graph) & $l/32$ \\
Hurricane (Jacobi set) & $l/64$ \\
\bottomrule
\end{tabu}
\label{tab:step-size}
\vspace{-4mm}
\end{table}

\para{Number of initial points for particle tracing.}
Particle tracing for contour extraction requires a set of starting points. In each span, MFA is a polynomial function and there could be multiple trajectories belonging to different parts of the same contour. We aim to find at least one starting point for each trajectory. We use gradient descent to locate the starting points (i.e.,~roots of polynomial functions) from a set of initial points. 
The contour at a fixed isovalue is a continuous curve that contains infinite number of points; in practice, we could only sample a finite set of starting points for trajectories within a span and connect them across spans.

For an MFA model, as the degree $p$ increases, the contours could become more complex. To account for this, we set the number of initial points to be $(p+3)^2$ for a polynomial of degree $p$. Specifically, in each dimension, we select $(p+3)$ initial points: two points placed on the boundary and $(p+1)$ points sampled uniformly. We assume that this sampling strategy allows us to find at least one starting point on each trajectory through gradient descent.

When extracting the Jacobi set $\mathbb{J}(f,g)$, we convert the problem to a contour extraction at isovalue $0$ of function $h$. The degree of $h$ depends on degrees of $f$ and $g$ and is given by $p_h=p_f+p_g-1$. Therefore, we set the number of initial points for Jacobi set extraction to be $(p_h+3)^2=(p_f+p_g+2)^2$.

Similarly, for ridge-valley graph extraction, we convert it to a contour extraction problem at isovalue 0 of $\Tilde{h}$. The degree of $\Tilde{h}$ is $p_{\Tilde{h}}=3p_f-1$. Consequently, we set the number of initial points to be $(p_{\Tilde{h}}+3)^2=(3p_f+2)^2$.

\para{Accuracy threshold.}~We control contour extraction accuracy via a tolerance $\epsilon$, so that points satisfy: $f(x)-a\leq \epsilon$. Both the Jacobi set and the ridge–valley graph can be reformulated as level $0$ contour extraction problems; their accuracies are similarly governed by tolerances $h\leq\epsilon$ and $\Tilde{h}\leq\epsilon$, respectively. To determine an appropriate value, we evaluate $\epsilon \in \{10^{-6},  10^{-8}, 10^{-10},  10^{-12}\}$ across all MFA models. According to the following ablation study, we find that $\epsilon = 1e^{-10}$ consistently provides sufficient precision, and thus use it for all our experiments. 

\para{Trajectory connection threshold.} 
We connect trajectories whose endpoints are within a connection threshold $\gamma$. 
Since the step size determines the spacing between points, it also influences the distance between trajectories. Therefore, we evaluate $\gamma \in \{s, 1.5s, 2.0s, 2.5s\}$ across all MFA models. Based on results from synthetic models, we find that $\gamma = 2s$ is a suitable choice for achieving correct connectivity.

\subsection{Ablation Study}
\begin{figure}[!ht]
    \includegraphics[width=\linewidth]{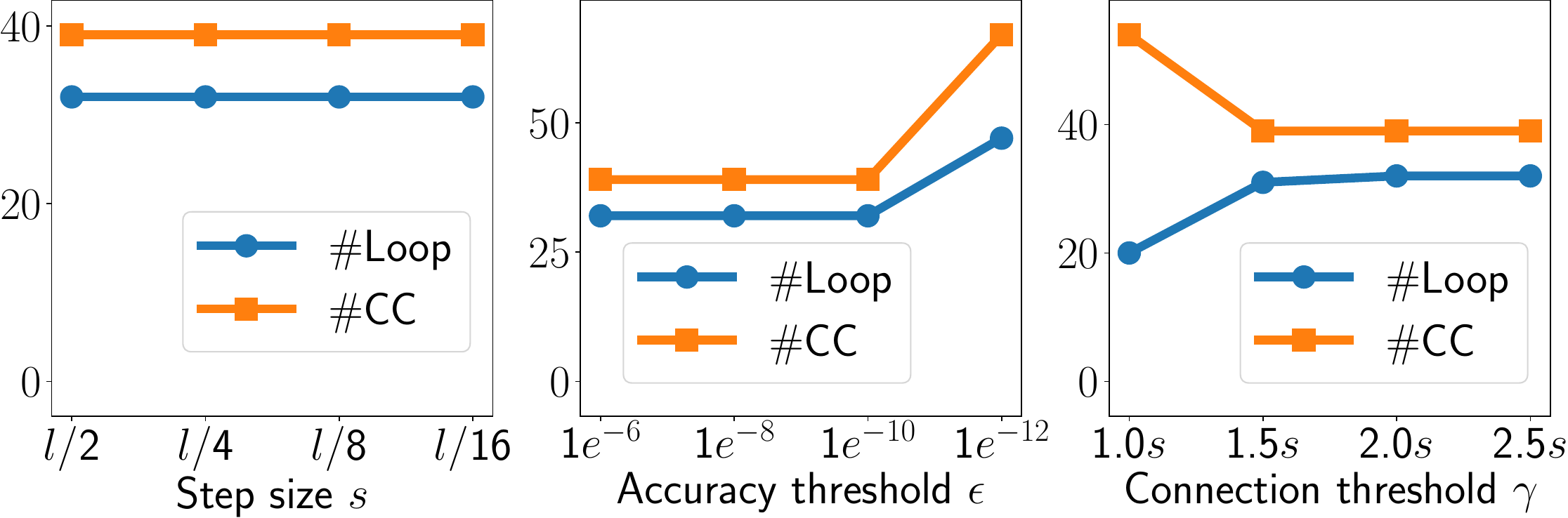}
    \vspace{-6mm}
    \caption{Schwefel model: Contour extraction at $a=100$.}
    \label{fig:schwefel-1-ablation-study}
\end{figure}

\begin{figure}[!ht]
    \includegraphics[width=\linewidth]{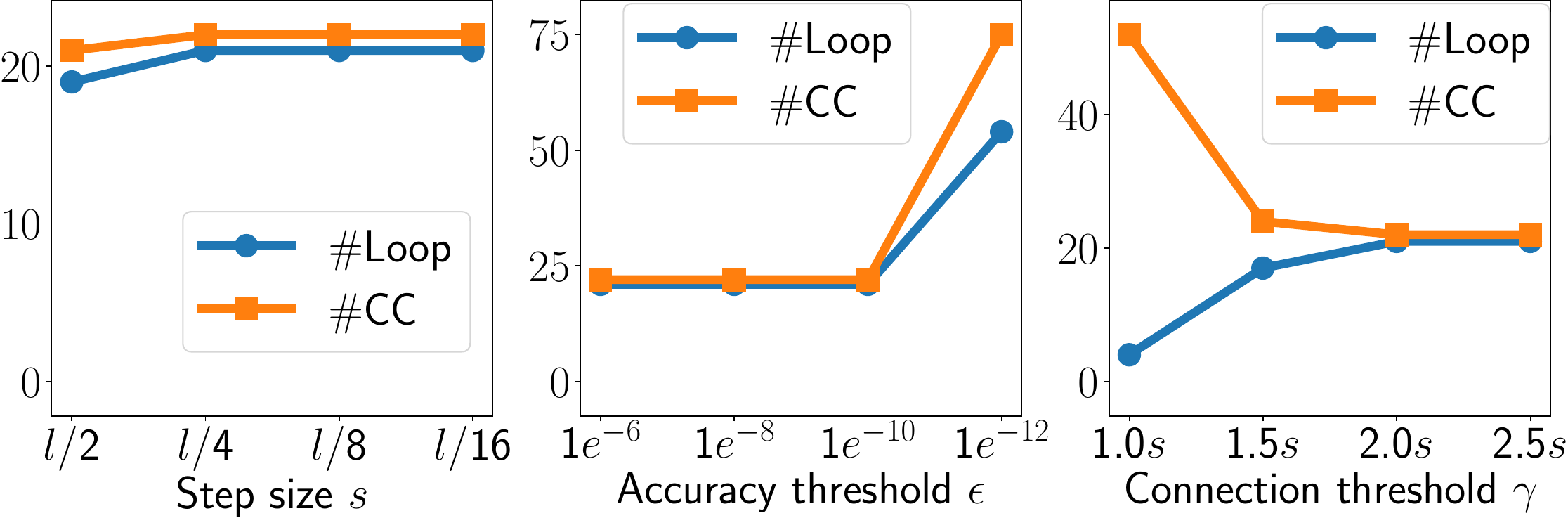}
    \vspace{-6mm}
    \caption{Schwefel model: Contour extraction at $a=500$.}
    \label{fig:schwefel-2-ablation-study}
    \vspace{-4mm}
\end{figure}

\para{Schwefel: contour extraction.}
We extract contours from the Schwefel MFA model at two isovalues, $a = 100$ and $500$ respectively. According to the convergence plots in \cref{fig:schwefel-1-ablation-study,fig:schwefel-2-ablation-study}, our method accurately recovers the ground truth when $s \leq l/4$. Fixing $s=l/4$, we further investigate the sensitivity of the contour extraction algorithm w.r.t. parameters $\epsilon$ and $\gamma$. 
We use $\epsilon = 1.0e^{-10}$ and $\gamma = 2.0s$ in our experiment based on these convergence plots.   

\begin{figure}[!ht]
    \includegraphics[width=\linewidth]{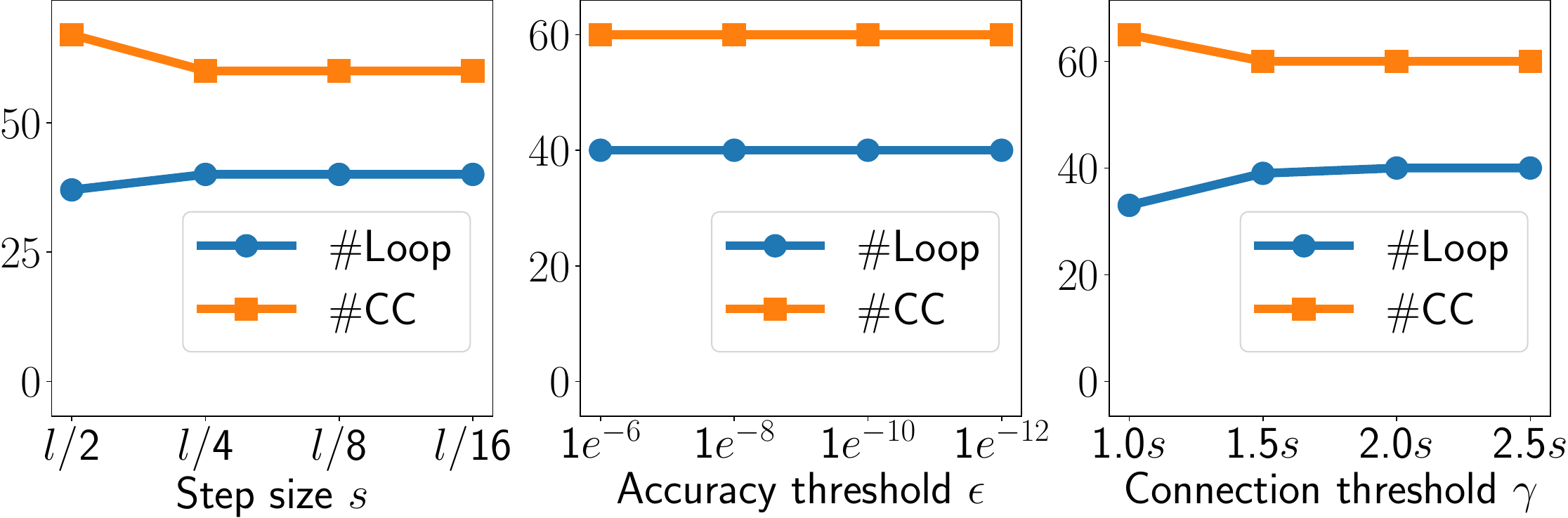}
    \vspace{-6mm}
    \caption{Sinc model: Contour extraction at $a=0.33$.}
    \label{fig:sinc-1-ablation-study}
    \vspace{-4mm}
\end{figure}

\begin{figure}[!ht]
    \includegraphics[width=\linewidth]{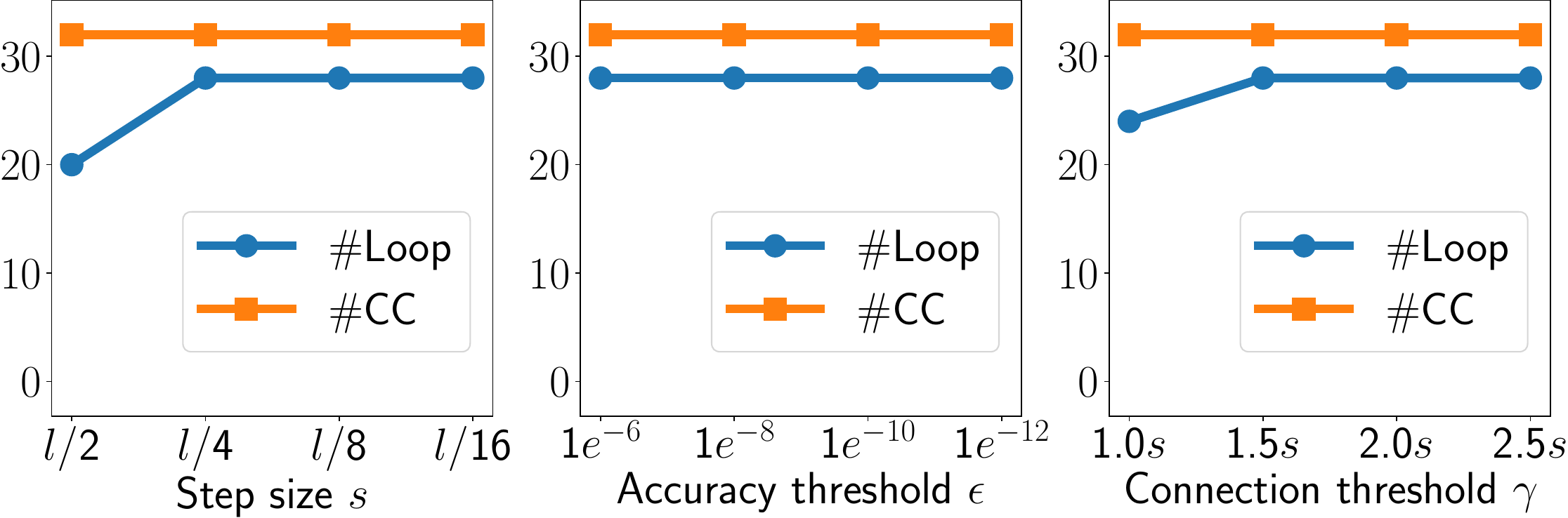}
    \vspace{-6mm}
    \caption{Sinc model: Contour extraction at $a=0.79$.}
    \label{fig:sinc-2-ablation-study}
    \vspace{-2mm}
\end{figure}

\para{Sinc: contour extraction.}
We perform contour extraction at different isovalues  $a=0.33$ (\cref{fig:sinc-1-ablation-study}) and $a=0.79$ (\cref{fig:sinc-2-ablation-study}). Based on the observed convergence behavior, we choose a step size of $s=l/4$. Additionally, we set $\epsilon = 1.0e^{-10}$ and $\gamma = 2.0s$ based on these convergence plots.

\begin{figure}[!ht]  
    \includegraphics[width=\linewidth]{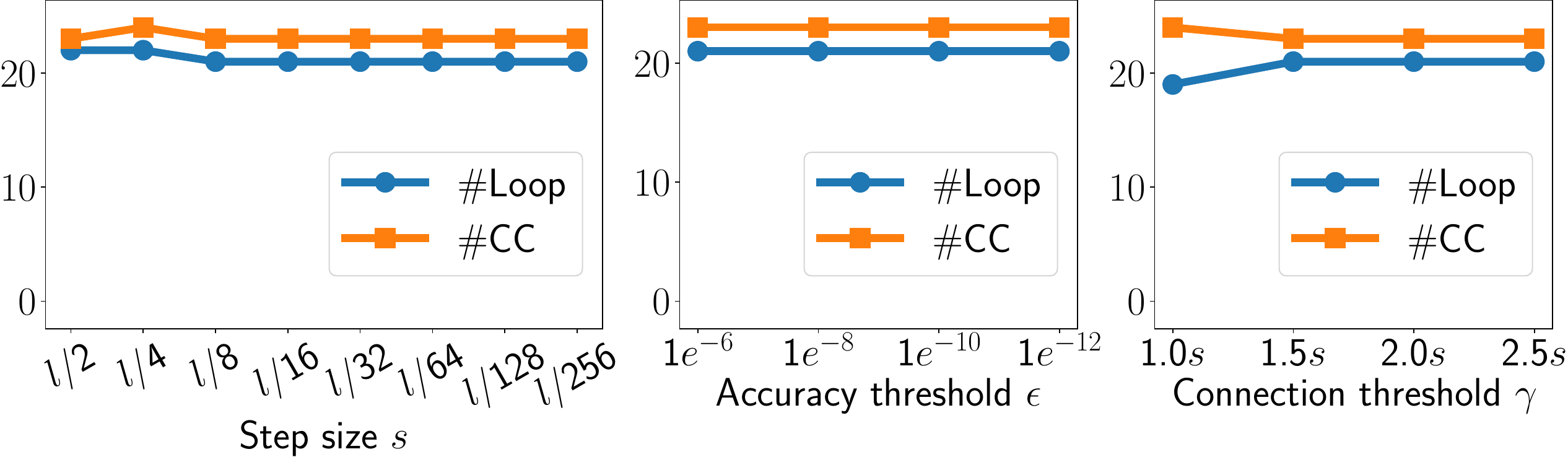}
    \vspace{-6mm}
    \caption{S3D model: Contour extraction at $a=30$.}
    \label{fig:s3d-1-ablation-study}
    \vspace{-2mm}
\end{figure}

\begin{figure}[!ht] 
    \includegraphics[width=\linewidth]{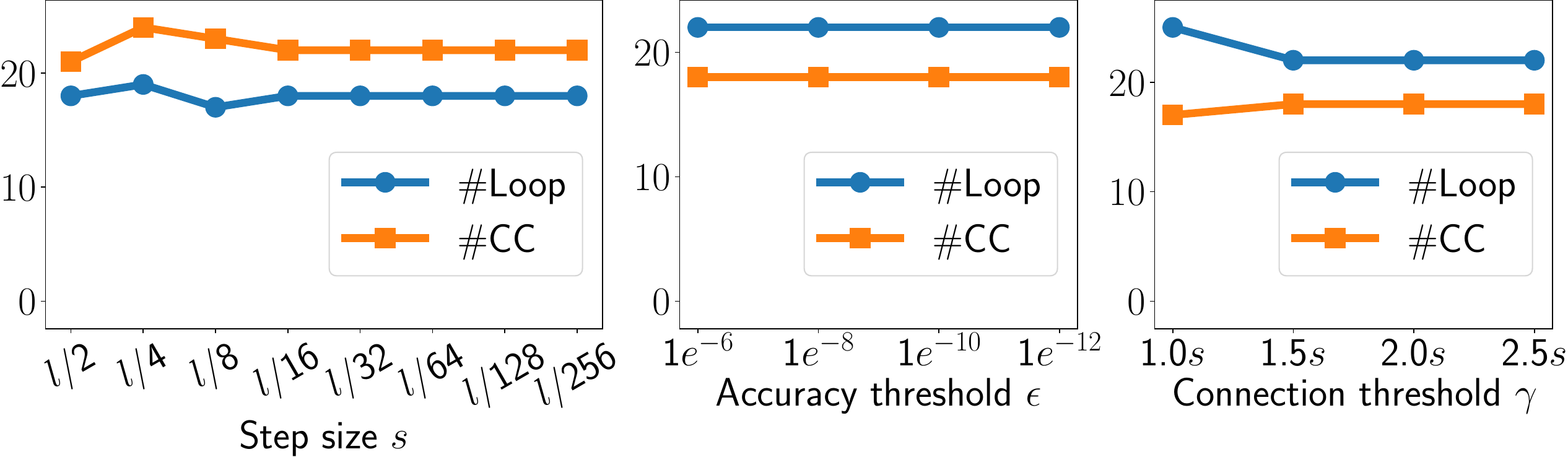}
    \vspace{-6mm}
    \caption{S3D model: Contour extraction at $a=50$.}
    \label{fig:s3d-2-ablation-study}
    \vspace{-2mm}
\end{figure}

\begin{figure}[!ht]  
    \includegraphics[width=\linewidth]{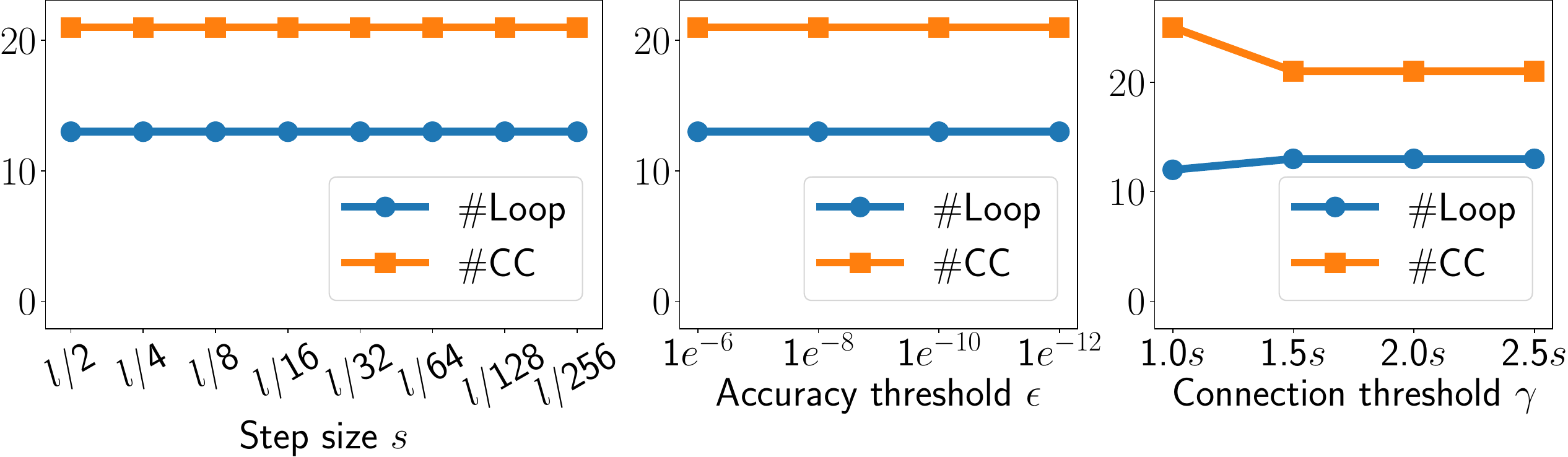}
    \vspace{-6mm}
    \caption{S3D model: Contour extraction at $a=60$.}
    \label{fig:s3d-3-ablation-study}
    \vspace{-2mm}
\end{figure}

\para{S3D: contour extraction.}
We perform contour extraction at three different isovalues $a = 30$ (\cref{fig:s3d-1-ablation-study}), $a = 50$ (\cref{fig:s3d-2-ablation-study}), and $a = 60$ (\cref{fig:s3d-3-ablation-study}). 
In these experiments, we adopt $s = l/16$, as both \#Loop and \#CC begin to stabilize at this point. 
These convergence plots also demonstrate the robustness of the accuracy threshold $\epsilon$. 
Consistent with all other experiments, we use $\epsilon = 1.0e^{-10}$ and $\gamma = 2.0s$.

\begin{figure}[!ht]
    \includegraphics[width=\linewidth]{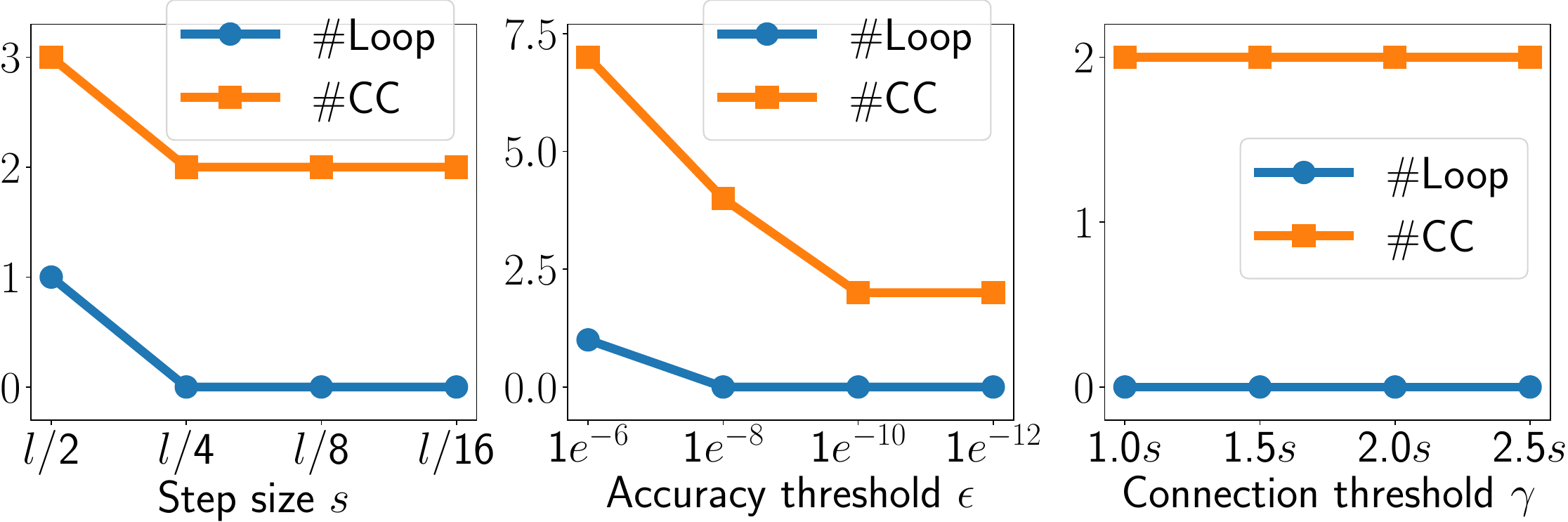}
    \vspace{-6mm}
    \caption{Gaussian pair model: Jacobi set extraction.}
    \label{fig:gaussian-pair-ablation-study}
    \vspace{-2mm}
\end{figure}

\para{Gaussian Pair: Jacobi set extraction.}
\cref{fig:gaussian-pair-ablation-study} displays convergence plots of \#Loop and \#CC for step size, showing convergence starting from $s = l/4$. 
These plots also show stable convergence for $\epsilon \leq 1.0\times10^{-10}$. 
Moreover, the plots demonstrate the robustness of extraction results w.r.t.~variations in the trajectory connection threshold $\gamma$.

\begin{figure}[!ht]
    \includegraphics[width=\linewidth]{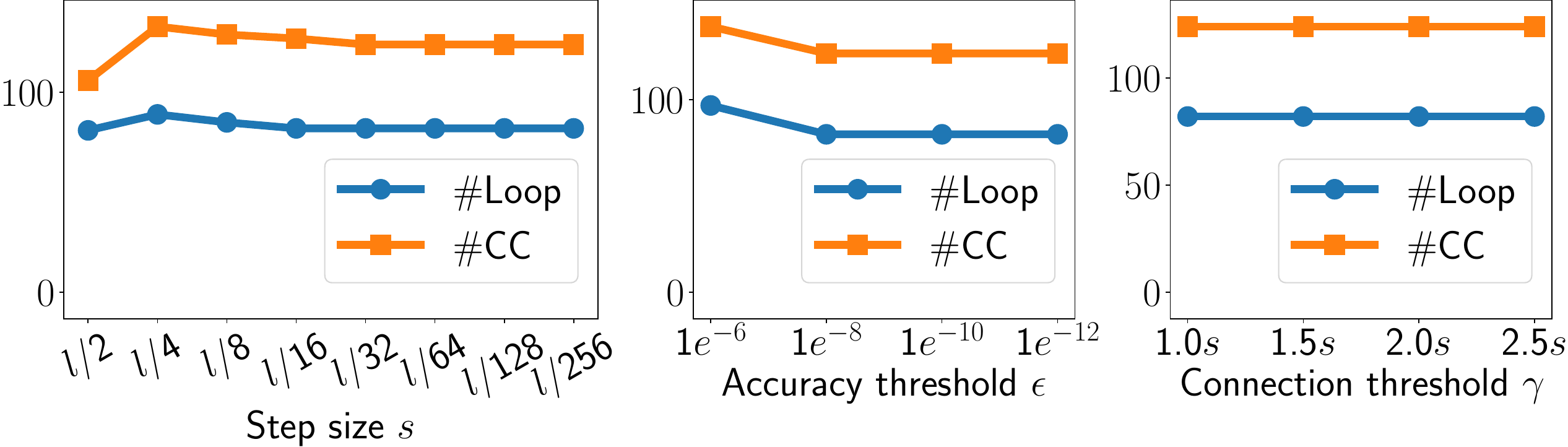}
    \vspace{-6mm}
    \caption{Von K\'arm\'an Vortex Street model: Jacobi set extraction.}
    \label{fig:karman-ablation-study}
    \vspace{-2mm}
\end{figure}

\para{Von K\'arm\'an Vortex Street: Jacobi set extraction.}
\cref{fig:karman-ablation-study} presents convergence plots for varying step sizes. 
We use $s = l/32$, marking the onset of convergence. 
Convergence remains stable for $\epsilon \leq 1.0\times10^{-8}$. 
Additionally, these plots demonstrate the robustness of extraction results against variations in the trajectory connection threshold $\gamma$.

\begin{figure}[!ht]
    \includegraphics[width=\linewidth]{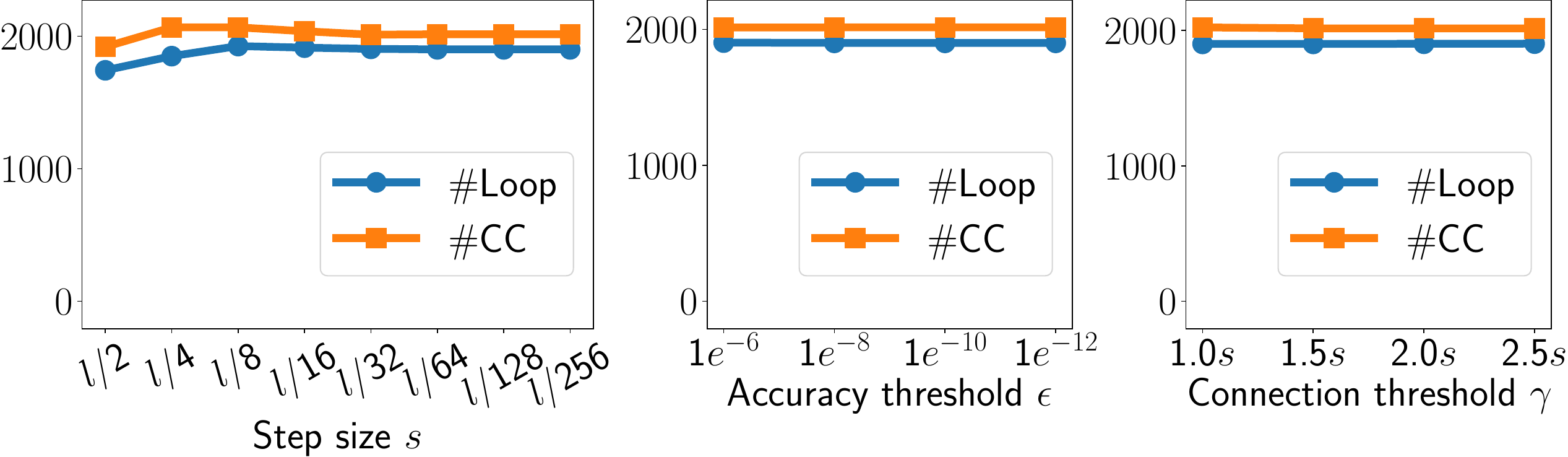}
    \vspace{-6mm}
    \caption{Hurricane Isabel model: Jacobi set extraction.}
    \label{fig:hurricane-ablation-study}
    \vspace{-2mm}
\end{figure}

\para{Hurricane Isabel: Jacobi set extraction.}
Similar to previous MFA models, convergence plots are shown in \cref{fig:hurricane-ablation-study}. The chosen step size is $s = l/64$.
These plots demonstrate that our extracted results are robust w.r.t. variations in $\epsilon$ and $\gamma$.

\begin{figure}[!ht]
        \includegraphics[width=\linewidth]{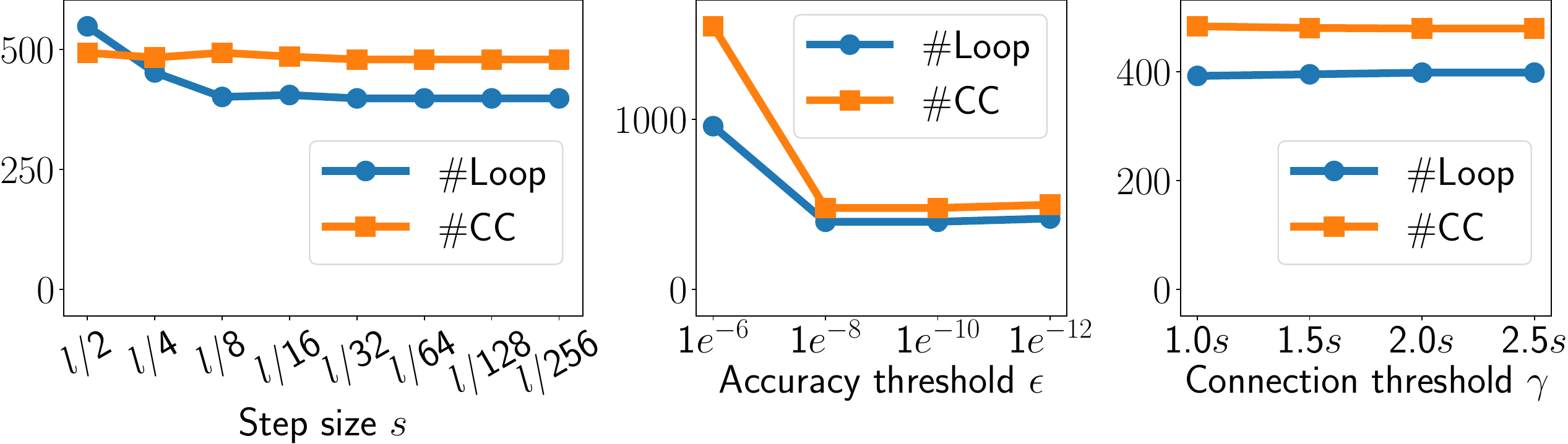}
                \vspace{-6mm}
        \caption{Boussinesq Approximation model: Jacobi set extraction.}
        \label{fig:boussinesq-ablation-study}
    \end{figure}

\para{Boussinesq Approximation: Jacobi set Extraction.}
\cref{fig:boussinesq-ablation-study} shows convergence plots of Jacobi set extraction with various step sizes. 
Our experiments use $s = l/32$, $\epsilon = 1.0e^{-10}$, and $\gamma = 2.0s$.

\begin{figure}[!ht]
    \includegraphics[width=\linewidth]{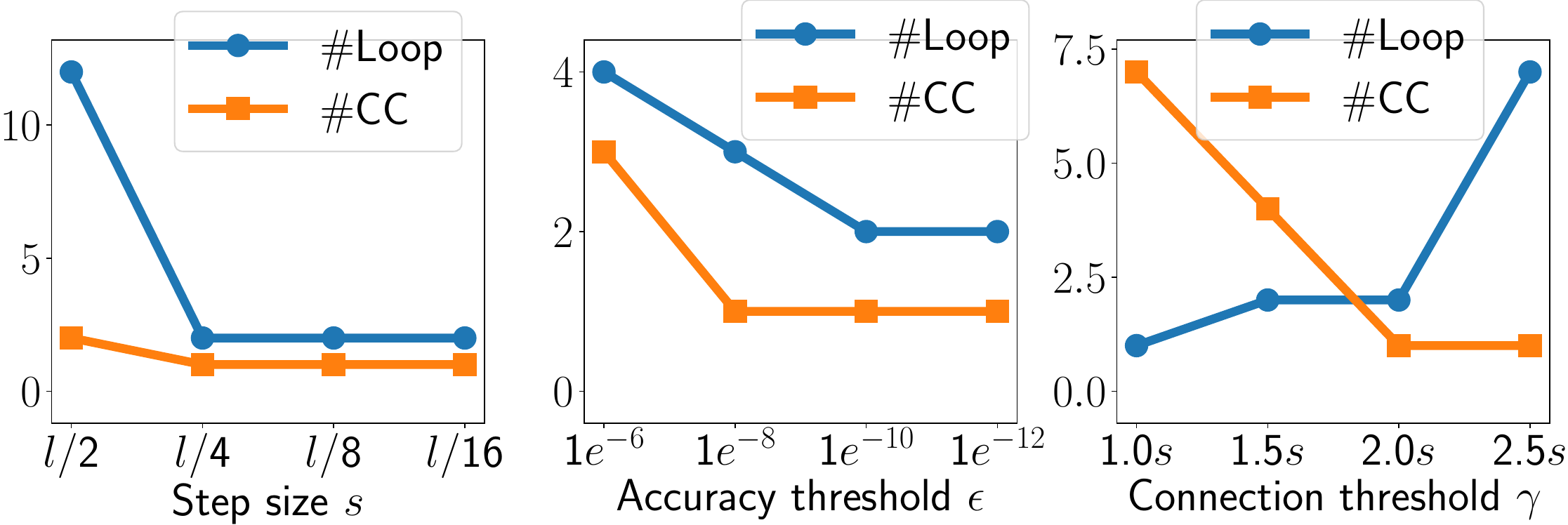}
    \vspace{-6mm}
    \caption{Gaussian mixture model: Ridge-valley graph extraction.}
    \label{fig:gaussian-mixture-ablation-study}
    \vspace{-2mm}
\end{figure}

\para{Gaussian Mixture: ridge-valley graph extraction.}
We illustrate convergence plots for ridge-valley graph extraction in \cref{fig:gaussian-mixture-ablation-study}. 
A step size of $s = l/4$ is employed. 
In this experiment, a large $\gamma$ may introduce additional loops; this effect is illustrated in the case of $\gamma = 2.5s$. We set $\epsilon = 1.0e^{-10}$ and $\gamma = 2.0s$ in our experiment.

\begin{figure}[!ht]
    \includegraphics[width=\linewidth]{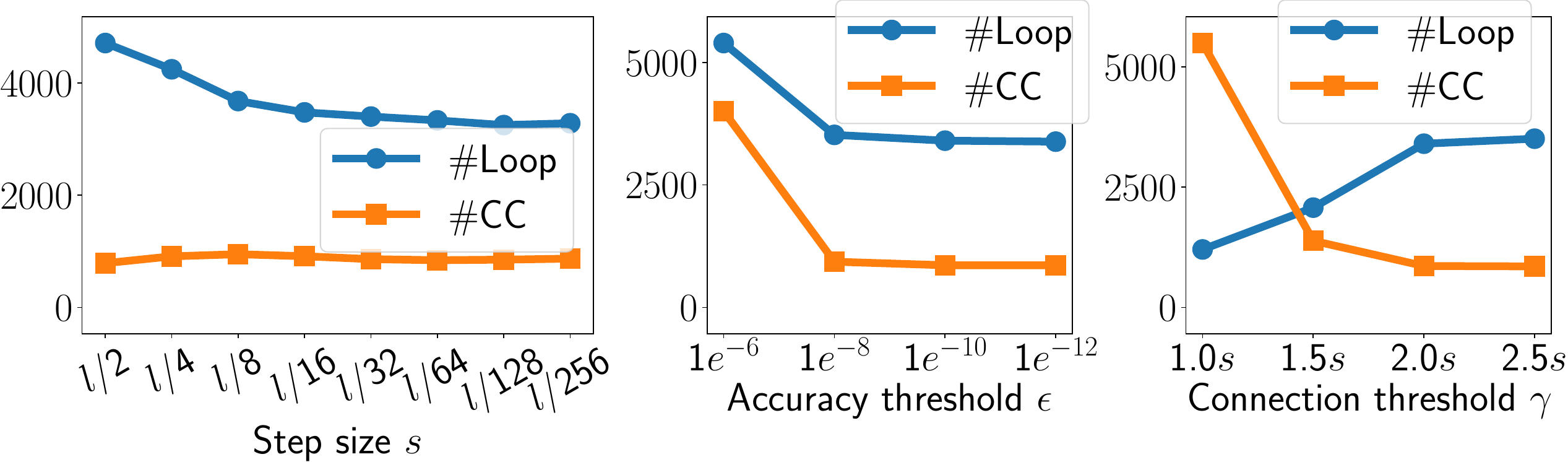}
    \vspace{-6mm}
    \caption{CESM model: Ridge-valley graph extraction.}
    \label{fig:cesm-ablation-study}
\end{figure}

\para{CESM: ridge-valley graph extraction.}
We show the convergence plots in \cref{fig:cesm-ablation-study}. The step size is chosen to be $s=l/32$. Additionally, we set $\epsilon = 1.0e^{-10}$ and $\gamma = 2.0s$.

\section{Additional Experimental Results}
\label{sec:additional-experiments}
\subsection{Schwefel: Contour Extraction}
\label{sec:schwefel-contour}
\begin{figure}[!ht]
\centering
\includegraphics[width=\linewidth]{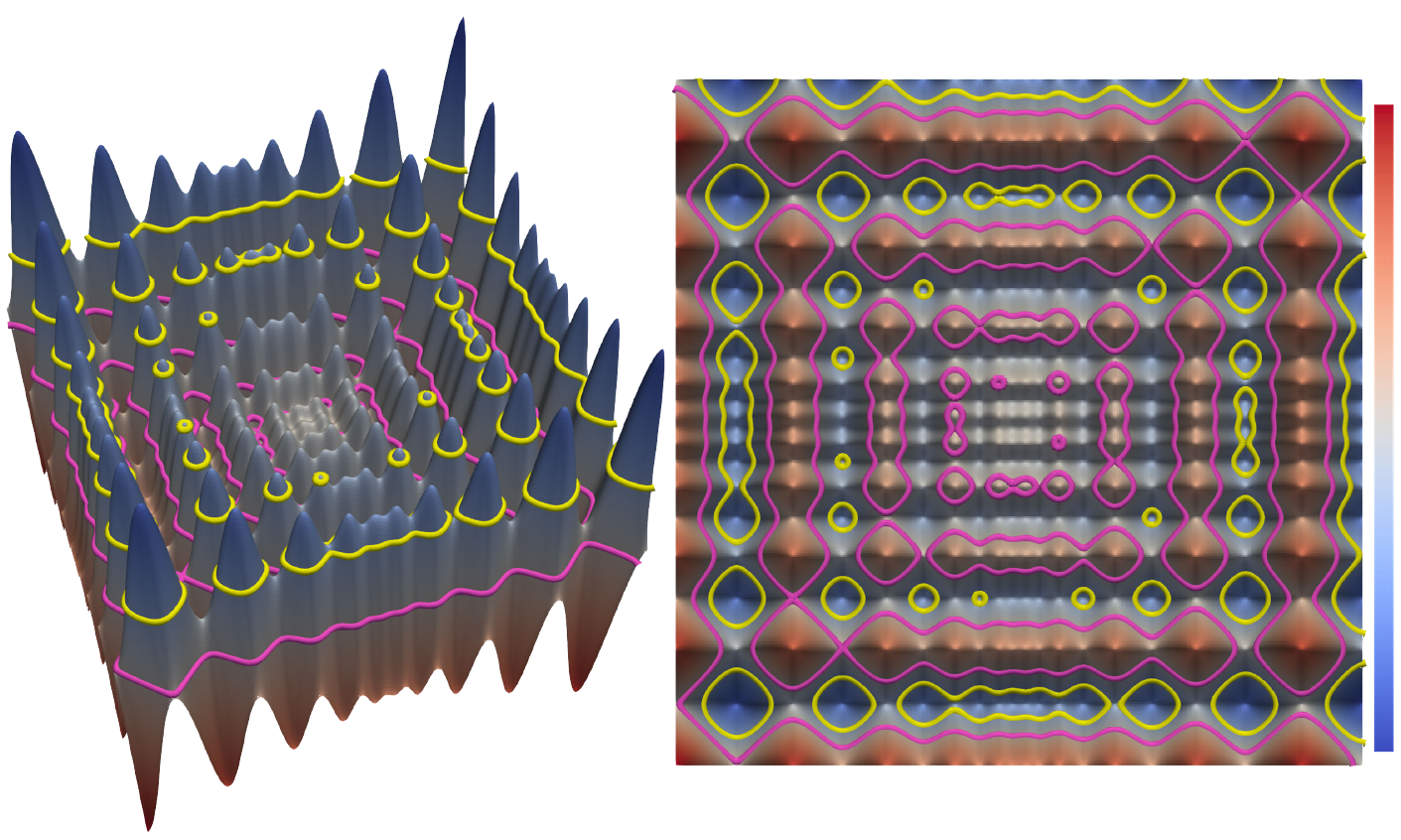}
\vspace{-7mm}
\caption{Schwefel model: contour extraction with isovalues $a = 100$ (yellow) and $a = 500$ (pink), shown from side and top viewpoints. }
\vspace{-3mm}
\label{fig:schwefel}
\end{figure}

As shown in \cref{fig:schwefel}, we extract contours from the Schwefel MFA model with isovalues $a=100$ and $a=500$. Since the closed-form function that the Schwefel model approximates is known, we use it as the ground truth. In \cref{tab:schwefel-contour}, all errors are below $\epsilon = 1e^{-10}$. Additionally, the number of loops and connected components matches that of the ground truth.

 \begin{table}[!ht]
 \caption{Schwefel model: Evaluation of contour extraction with various isovalues ($a$). GT denotes the ground truth.}
\scriptsize
\centering
\begin{tabu}{c|cccccc}
\toprule
$a$ & $\emax$ & $\eavg$ & \#Loop & GT\#Loop  & \#CC  & GT\#CC\\ 
\midrule
100 & $9.9e^{-11}$ & $3.5e^{-12}$ & 32 & 32 & 39 & 39 \\
500 & $1.0e^{-10}$ & $3.5e^{-12}$ & 21 & 21 & 22 & 22\\
\bottomrule
\end{tabu}
\label{tab:schwefel-contour}
\vspace{-6mm}
\end{table}

\subsection{Boussinesq Approximation: Jacobi Set Extraction}

\begin{figure}[!ht]
\centering
\includegraphics[width=\linewidth]{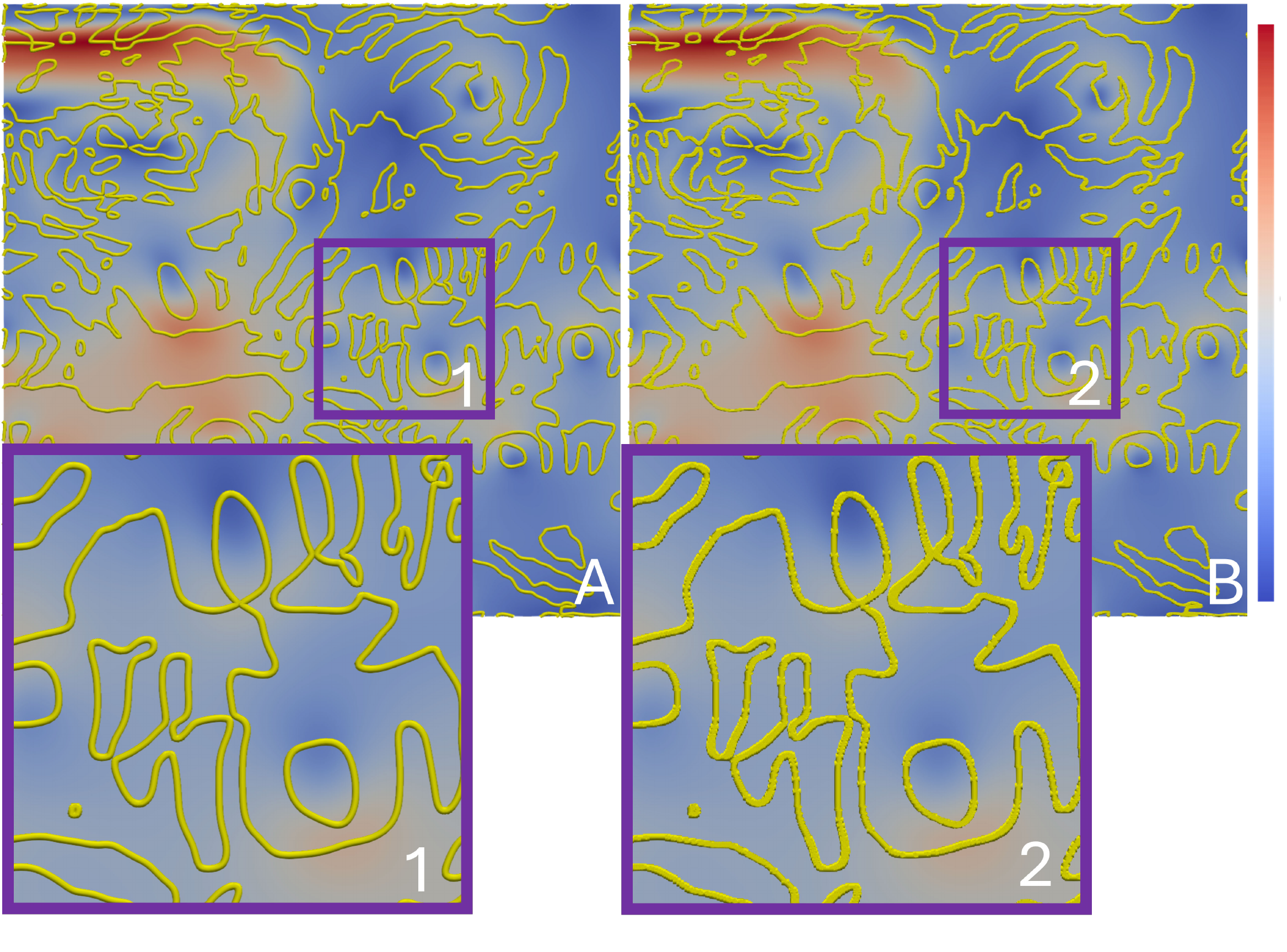}
\vspace{-7mm}
\caption{Boussinesq approximation model: Jacobi set extraction using MFA continuous method (A) and discrete method (B). (1) and (2): zoomed-in views of the purple blocks in (A) and (B), respectively.}
\label{fig:boussinesq-approximation}
\end{figure}

In \cref{fig:boussinesq-approximation}, we display the Boussinesq approximation model at time step 2000, with the extracted Jacobi set highlighted in yellow using step size $s=l/32$. Because our framework operates on a continuous representation, the results remain smooth; see blocks (1). In contrast, the discrete method produces zigzag patterns; see block (2).  
The evaluation metrics in~\cref{tab:boussinesq-jacobi-set} indicate that the discrete method generates a large number of spurious loops. 

\begin{table}[!ht]
\vspace{-2mm}
\caption{Boussinesq approximation model: Jacobi set extraction of MFA models with step size $s=1/32$ and sampling ratio 32.} 
\scriptsize
\centering
\setlength{\tabcolsep}{3pt} 
\begin{tabu}{cccc|cccc}
\toprule
\multicolumn{4}{c|}{MFA Continuous Method} &  \multicolumn{4}{c}{Discrete Method}\\
\midrule 
$\emax$ & $\eavg$ &\#Loop & \#CC & $\emax$ & $\eavg$ &\#Loop & \#CC \\ 
\midrule
$1.0e^{-10}$& $1.6e^{-11}$ & 398 & 479 & 10.3 & $2.7e^{-2}$ & 43888  & 9651 \\
\bottomrule
\end{tabu}
\label{tab:boussinesq-jacobi-set}
\vspace{-4mm}
\end{table}

\end{document}